\pdfoutput=1
\documentclass{article}

\usepackage[utf8]{inputenc} 
\usepackage[T1]{fontenc}    
\usepackage{hyperref}       
\usepackage{url}            
\usepackage[round]{natbib}
\usepackage{booktabs}       
\usepackage{amsfonts}       
\usepackage{nicefrac}       
\usepackage{microtype}      
\usepackage{amsmath}
\usepackage{ulem}  
\usepackage{comment} 
\usepackage{multirow}
\usepackage{graphicx}
\usepackage{float}
\usepackage{subfig}
\usepackage{mathtools}
\usepackage{wrapfig}
\usepackage{array}
\newcolumntype{?}{!{\vrule width 1.5pt}}
\usepackage{amsthm}
\theoremstyle{definition}
\newtheorem{defn}{Definition}[section]

\begin{document}

\title{Estimating a Manifold from a Tangent Bundle Learner}

\author{Bharathkumar Ramachandra\thanks{Corresponding author.} \footnotemark[2], Benjamin Dutton\thanks{Authors contributed equally.} \\ and Ranga Raju Vatsavai\\ 
Department of Computer Science, \\ North Carolina State University\\
Raleigh, North Carolina 27695-8206\\
Email: bramach2@ncsu.edu, bcdutton@ncsu.edu \\and rrvatsav@ncsu.edu}

\date{}

\maketitle


\begin{abstract}
Manifold hypotheses are typically used for tasks such as dimensionality reduction, interpolation, or improving classification performance. In the less common problem of manifold estimation, the task is to characterize the geometric structure of the manifold in the original ambient space from a sample. We focus on the role that tangent bundle learners (TBL) can play in estimating the underlying manifold from which data is assumed to be sampled. Since the unbounded tangent spaces natively represent a poor manifold estimate, the problem reduces to one of estimating regions in the tangent space where it acts as a relatively faithful linear approximator to the surface of the manifold.  
Local PCA methods, such as the Mixtures of Probabilistic Principal Component Analyzers method of Tipping and Bishop produce a subset of the tangent bundle of the manifold along with an assignment function that assigns points in the training data used by the TBL to elements of the estimated tangent bundle. We formulate three methods that use the data assigned to each tangent space to estimate the underlying bounded subspaces for which the tangent space is a faithful estimate of the manifold and offer thoughts on how this perspective is theoretically grounded in the manifold assumption. We seek to explore the conceptual and technical challenges that arise in trying to utilize simple TBL methods to arrive at reliable estimates of the underlying manifold.
\end{abstract}

\section{Introduction}
\label{sect:intro}
\begin{figure}[ht]
\centering
\includegraphics[trim={4cm 3cm 3cm 4.5cm}, clip, scale=0.6]{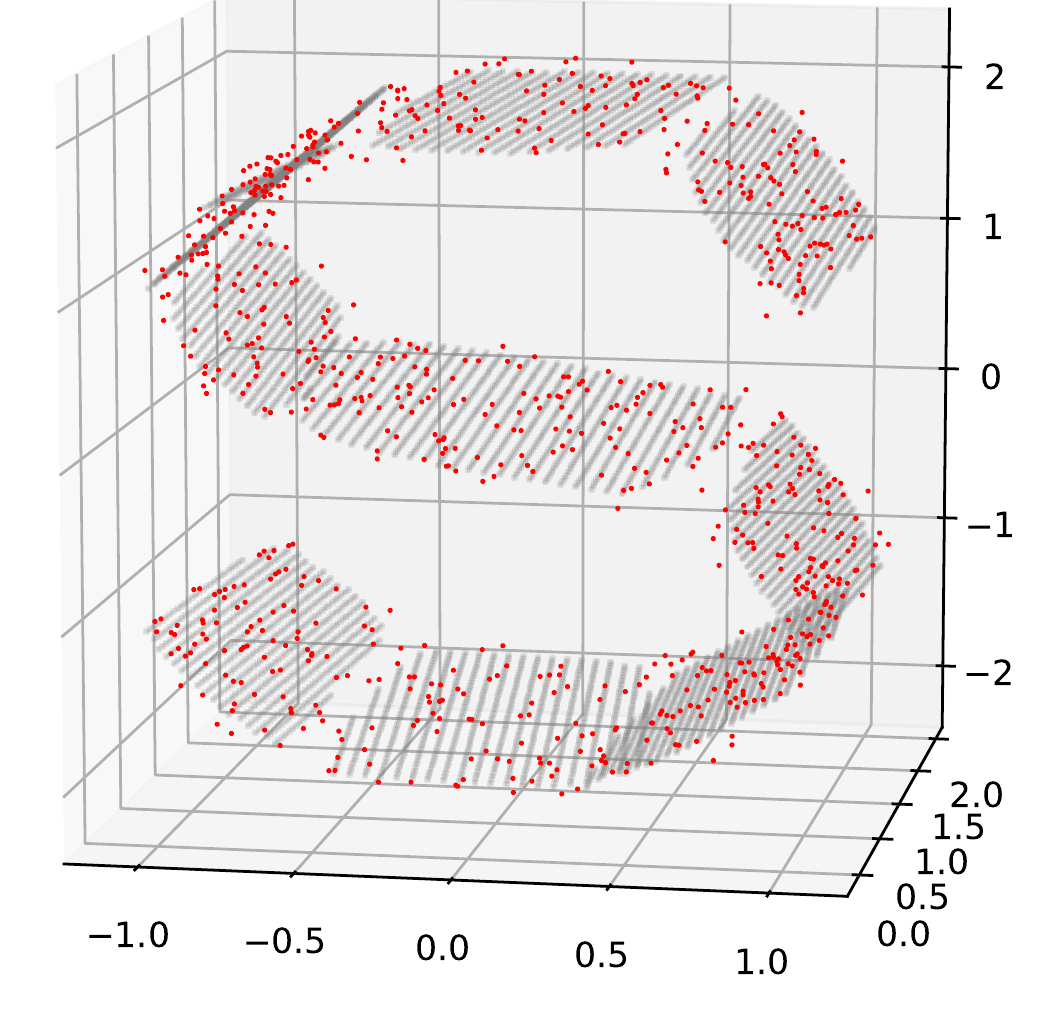}
\caption{An estimate of the manifold of an S-curve in gray, from a set of training points in red.}
\label{fig:catch}
\end{figure}

Manifolds are topological spaces that have a property of local flatness. That is, they locally resemble Euclidean space in some neighborhood around each point on the manifold. Formally, each point on an $d$-dimensional manifold has a neighborhood that is homeomorphic to $d$-dimensional Euclidean space. Learning from high dimensional data is a challenging problem due to the curse of dimensionality, and manifold learning methods operate under the assumption of one or more manifold hypothesis to circumvent this curse.  These assumptions allow the manifold structure of the data to be exploited in various ways. The manifold hypothesis for unsupervised learning, for instance, is that real-world high dimensional data lie on or near a manifold of much lower intrinsic dimension \citep{cayton2005algorithms,narayanan2010sample}. Various learning tasks such as dimensionality reduction, on-manifold interpolation, manifold-aware clustering and manifold-aware density estimation have been explored under this hypothesis. 

One important task that has received less attention is manifold estimation, which aims to characterize the learned manifold as a subset of the ambient space (see Figure \ref{fig:catch}). 
Other tasks such as manifold-aware classification and manifold-aware semi-supervised methods fall under alternate manifold hypotheses for supervised and semi-supervised learning respectively \citep{rifai_manifold_2011}. 
Existing methods that perform manifold estimation \citep{cheng_manifold_2005,boissonnat_manifold_2009,boissonnat_manifold_2014} have primarily been explored in the specific cases of 1D or 2D manifolds embedded in $\mathbb{R}^2$ and $\mathbb{R}^3$. These computational geometric approaches are impractical for real-world high dimensional data because they assume noise-free sampling from the manifold or use the weighted Delaunay triangulation, which is infeasible in high dimensions. 

In this work, we formulate three methods that perform manifold estimation using an estimated tangent bundle and a relation that associates points in training data with tangent spaces in the tangent bundle.  For each tangent space, we use points in training data for which the tangent space was a good linear approximator and seek to generalize to entire neighborhoods.  We call this problem Faithful Neigborhood Estimation (FNE). Once estimated, the faithful neighborhoods can approximate the manifold by simply considering the union of these neighborhoods. We construct three FNE techniques and evaluate the quality of our estimates on synthetic 2D and 3D benchmark datasets for manifold learning. In addition, we study the variation of evaluation metrics as the training set size $n$ and the ratio of training set size to the number of tangent spaces $n/k$ varies. 


\section{Mathematical Preliminaries and Problem Definition}
A \textbf{homeomorphism} is a continuous function with a continuous inverse.  In topology, homeomorphisms are isomorphisms.  In other words, they are functions that preserve topological structure.  A \textbf{d-dimensional manifold} $\mathcal{M}$ is a set that is locally homeomorphic to $\mathbb{R}^d$.  Each point $x \in \mathcal{M}$ belongs to at least one open neighborhood $\mathcal{N}_{i}$ that can be mapped via a homeomorphism $C_{i}: \mathcal{N}_{i} \rightarrow \mathbb{R}^d$.  These mappings are called \textbf{coordinate charts} (or just \textbf{charts}) because they assign local coordinates to points in their domain. 

Natively, manifolds are purely topological objects, but here we consider only manifolds in $\mathbb{R}^D$ because we assume a data sample $\mathcal{X} \subset \mathbb{R}^D$ from which we seek to estimate $\mathcal{M}$ with $\hat{\mathcal{M}}$.  In addition, we assume $\mathcal{M}$ is differentiable. A manifold is differentiable if, for every atlas in the equivalence class for that manifold, the transition charts are differentiable.  A \textbf{transition chart} $\tau_{i,j}: C_i(\mathcal{N}_i \cap \mathcal{N}_j) \rightarrow C_j(\mathcal{N}_i \cap \mathcal{N}_j) \coloneqq C_j(C_i^{-1}(\cdot))$ gives a way of comparing coordinates assigned by two charts defined over the same region.  
    
The collection of all tangent vectors at a point $x$ on a differentiable manifold $\mathcal{M}$ form the \textbf{tangent space} of $\mathcal{M}$ at $x$, which we denote as $T_x\mathcal{M}$ or simply $T_x$ when it is clear which manifold we are referring to.  The tangent space of a manifold has the same dimension as the manifold: $d$.  The collection of all tangent spaces for $\mathcal{M}$, together with the points at which they are the derivative (which we call \textit{anchor points}),  $\big\{ (x, T_x\mathcal{M}) \big\})_{x\in \mathcal{M}}$ is called the \textbf{tangent bundle} of $\mathcal{M}$, which we denote as $TB(\mathcal{M})$.
    
    
    
\begin{defn}[Manifold Estimation in $\mathbb{R}^D$]
Given a noisy sample of $n$ points $\mathcal{X} \subset \mathbb{R}^D$ that lie on or near a $d$-dimensional manifold $\mathcal{M} \subset \mathbb{R}^D$, the \textbf{manifold estimation} task is to estimate $\mathcal{M}$ with $\hat{\mathcal{M}} \subset \mathbb{R}^D$ using $\mathcal{X}$.
\end{defn}

The $D$ and $d$ mentioned above will henceforth be referred to as the \textbf{ambient} dimension and \textbf{intrinsic/latent} dimension as is convention.

This task should not be confused with the more common and related task of \textit{manifold embedding}, where a low-dimensional parameterization of $\mathcal{X} \subset \mathcal{M}$ or, ideally, $\mathcal{M}$ itself is sought. Acknowledging that the manifold itself lives in ambient space, manifold estimation seeks only a subset of the ambient space, $\mathbb{R}^D$, that suitably characterizes the structure.

Usually, manifold learning tasks all share a common first step in estimating the intrinsic dimensionality of the manifold, and hence this has been relatively well-researched. The reader is referred to \citep{verveer_evaluation_1995,levina_maximum_2005} for relevant research in this field. In dealing with the manifold estimation task, we assume that intrinsic dimension has been previously determined, and refer to it as either $d$ or $\hat{d}$.

\begin{defn}[Tangent Bundle Learning in $\mathbb{R}^D$]
Given a noisy sample of $n$ points $\mathcal{X} \subset \mathbb{R}^D$ that lie on or near a $d$-dimensional manifold $\mathcal{M} \subset \mathbb{R}^D$ along with an estimate of its dimensionality, $\hat{d}$ the \textbf{tangent bundle learning} task is to estimate the tangent bundle of $\mathcal{M}$, $TB(\mathcal{M}) \coloneqq \big\{ (x, T_x\mathcal{M})\big\}_{x \in \mathcal{M}}$ with $\hat{TB}(\mathcal{M}) = \big\{ (a_i, T_{a_i}\mathcal{M}) \big\}_{i=1}^{k} \subset TB(\mathcal{M})$ where $a_i \in \mathcal{M}$ but not necessarily $\in \mathcal{X}$, each of the tangent spaces $T_{a_i}\mathcal{M}$ are $\hat{d}$ dimensional, and $k$ is the total number of tangent spaces returned. 
\end{defn}

It is common for $k$ to be an input to a TBL method.  Additionally, TBL methods also usually return a relation (or function) that associates points in $\mathcal{X}$ with one or more tangent spaces.  The association can be based on many things, but we assume it associates points in $\mathcal{X}$ to tangent spaces in $\hat{TB}(\mathcal{M})$ which were good linear approximators for the points.  We refer to this relation as the \textit{tangent space assignment relation} (or just \textit{assignment relation}, henceforth).

\begin{defn}[Tangent Space Assignment Relation]
A \textbf{tangent space assignment relation} $A \subset \mathcal{X} \times \hat{TB}(\mathcal{M})$ associates tangent spaces in $\hat{TB}(\mathcal{X})$ to points in $\mathcal{X}$ for which they were a good linear approximator.
\end{defn}

Let $A_x$ be the set of tangent spaces associated with $x \in \mathcal{X}$ and $A_{T_y}$ be the set of points in $\mathcal{X}$ associated with $T_y$.  This relation is essential for the work here for the following reason. If a tangent space $T_y$ was a good linear approximator for $A_{T_y}$, then we seek to estimate the full neighborhood $N_y$, with $y \in N_y$ for which $A_y$ is a good linear approximator to $\mathcal{M}$.  We call this problem \textit{faithful neigborhood estimation}.


\begin{defn}[Faithful Neighborhood Estimation]
For all $(y, T_y\mathcal{M}) \in \hat{TB}(\mathcal{M})$ let $N_y$ be the set of points for which $T_y\mathcal{M}$ is a good linear approximator to $\mathcal{M}$.  The task of estimating $N_y$ using $y$, $T_y\mathcal{M}$, and $A_{T_y\mathcal{M}}$ is called \textbf{faithful neigborhood estimation} (FNE).
\end{defn}


\begin{defn}[Manifold Estimation from a Tangent Bundle Learner Problem]
Given a sample $\mathcal{X}$ from on or near $\mathcal{M}$, together with the estimated tangent bundle $\hat{TB}(\mathcal{M})$ produced from $\mathcal{X}$ by a TBL and an assignment relation $A$, the \textbf{manifold estimation from a tangent bundle learner problem} is to estimate $\mathcal{M}$ using $\hat{TB}(\mathcal{M})$, $\mathcal{X}$ and $A$.
\end{defn}

In this work, we explore three possible FNE algorithms that estimate $N_y$ from $\mathcal{X}_{T_y{\mathcal{M}}}$, which we denote M1, M2 and M3.  We argue that a reliable manifold estimate, $\hat{\mathcal{M}}$ can be found by simply considering the union of all the estimated neighborhoods.  That is, we argue that 
\begin{equation}
\hat{\mathcal{M}} \coloneqq \cup_{y \in \hat{TB}(\mathcal{M})} N_y \approx \mathcal{M}
\label{eqn:M-hat}
\end{equation}
is a viable manifold estimate and makes use of existing TBL algorithms toward this aim.

If we na\"ively assume $N_y = T_y\mathcal{M}$, then from Equation \ref{eqn:M-hat}, $\hat{\mathcal{M}}$ looks something like Figure \ref{fig:naive-lpca} because $T_y\mathcal{M}$ is an unbounded superset of $N_y$ and the Hausdorff distance between $\hat{\mathcal{M}}$ and $\mathcal{M}$ is infinite.  Utilizing $X_{T_y\mathcal{M}}$ and $T_y\mathcal{M}$ to estimate $N_y$ reduces to the task of estimating the region in $T_y\mathcal{M}$ where it acts as a relatively faithful estimate of $\mathcal{M}$, which is what methods M1, M2 and M3 explore.
    

\section{Related Work}
\subsection{Tangent Bundle Learners}
Data representation with low dimensional affine subspaces has been a heavy topic of research in the manifold learning community, mainly because it follows naturally from the locally flat property of manifolds. Many methods work under the observation that the least squares minimizing hyperplane for a flat neighborhood is spanned by the largest eigenvectors returned from Principal Components Analysis. Tangent bundle learners approximate the tangent bundle of the true manifold with a finite number of tangent hyperplanes.

To learn a tangent bundle, one approach is to alternate between steps of flat neighborhood estimation and subspace estimation while either maximizing likelihood of the data under a probabilistic model (such as \citep{tipping_mixtures_1999}) or minimizing reconstruction errors (such as \citep{zhang2009median, cappelli_multispace_2001}). \cite{fukunaga_algorithm_1971} formulated the first method that performed local tangent space estimation that later gave rise to a whole family of local manifold learning algorithms called Local Principal Components Analysis (Local PCA or LPCA henceforth). For this family of methods, the tangent spaces correspond to the spans of the principal eigenvectors from local covariance matrices and global coordination usually involves some way of determining the anchor points for the tangent spaces and a way to orient the tangent spaces to obtain a single, globally valid coordinate system that agrees with each individual one. \cite{kambhatla1997dimension} present VQPCA which uses the generalized Lloyd algorithm to iteratively perform steps of vector quantization and local PCA while minimizing reconstruction error between training data points and their reconstructions from the PCA projection. \cite{karygianni2014tangent} learn a tangent bundle from a constrained agglomerative clustering problem formulation for manifold samples with a difference of tangents criterion. The constraints restrict the tangent bundle to satisfy geometric properties of manifolds using the neighborhood graph of the data sample.

A number of methods such as in \citep{hinton_recognizing_1995,bregler1995nonlinear,ghahramani1996algorithm,tipping_mixtures_1999,brand_charting_2003,vincent_manifold_2003} all perform a ``mixtures of pancakes'' type of estimation where the pancakes are flattened out Gaussians, reflected in their covariance structures. The tangent spaces where these Gaussians lie, together form the learned tangent bundle. \cite{hinton_recognizing_1995} use k-means followed by PCA in each cluster (neighborhood) via linear auto-encoders to estimate a tangent bundle. Similarly, \cite{bregler1995nonlinear} arrive at a tangent bundle by performing k-means clustering to determine neighborhoods, performing PCA in each of these patches and fine tuning using an expectation-maximization (EM) procedure. An exact EM algorithm for estimating the parameters of a mixture of factor analyzers was formulated in \citep{ghahramani1996algorithm}, which can be thought of as a reduced dimension mixture of Gaussians. This also provides an estimate of density of points in ambient space \citep{tipping_mixtures_1999}.

Charting was proposed in \citep{brand_charting_2003} and is performed by expressing a mapping as a kernel-based mixture of linear projections. They formulate a posterior with standard Gaussian Mixture Model (GMM) likelihood function while simultaneously penalizing uncertainty due to inconsistent projections in the mixture. Charting provides in addition to an embedding, both the forward and reverse mappings between ambient and low-dimensional coordinate spaces, and density estimates in these spaces. \cite{vincent_manifold_2003} formulate a manifold-aware version of the Parzen windows density estimator that places pancaked Gaussians around each point instead of spherical ones. The tangent spaces from SVD of neighborhood covariance matrix around each point could be treated as a tangent bundle. \cite{lui2011tangent} factorize data tensors using a modified higher order Singular Value Decomposition (SVD) to a fixed set of tangent spaces (tangent bundle) on a Grassmann manifold and use the logarithmic map of the Grassmanian to get tangent vectors and use the canonical Riemannian metric to measure lengths of these tangent vectors which they use as a distance measure for nearest-neighbor type classification. \cite{rifai_learning_2011} extract tangent spaces local to each training point from the SVD of the Jacobian matrix of the encoder function of a Contractive Auto-Encoder (CAE) (formalized in \citep{rifai_manifold_2011}). The tangent spaces along with the training data points they correspond to form a tangent bundle estimate. 
In \cite{tyagi2013tangent}, the authors measure the estimation error between estimated tangent spaces and true tangent spaces using the angle between them, which could be generalized to evaluate tangent bundle learners.

\subsection{Manifold Estimation}
In manifold estimation, the problem is to find a (usually compact) set in ambient space that represents the low-dimensional manifold structure in a suitable sense. Usually this involves assuming that the manifold is differentiable such that it is meaningful to reason about tangent spaces to the manifold and compactness of the manifold. When the setting is in specific of a 2 dimensional manifold embedded in 3 dimensions, the problem is called surface reconstruction and has attracted much research from the computer graphics, computer vision and computational geometry communities.

The first breakthroughs in extending surface reconstruction to arbitrary dimensions came from \cite{cheng_manifold_2005} and \cite{boissonnat_manifold_2009}. Although they provide guarantees on quality of approximation, these methods are impractical for two reasons. First, they assume very dense, noise-free samples, which does not extend to real-world datasets. Second, they use weighted Delaunay triangulation in ambient dimension, which is infeasible for high dimensional data.  
Additionally, they assume a connected manifold, which is not a property that data manifolds necessarily possess. In fact, manifold hypotheses based on the existence of sub-manifolds often implicitly assume the opposite. Further, the methods that seek global coordination of local tangent spaces also assume connectedness of the underlying manifold.

\cite{boissonnat_manifold_2014} present the first algorithm which depends linearly on the ambient space dimension, quadratically on sample size and exponentially on intrinsic dimension. They construct a simplicial complex based on approximations to the tangent bundle of the manifold \citep{freedman_efficient_2002}. However, their methods are still practical only for small intrinsic dimensions due to weighted Delaunay triangulation.

Manifold estimation also has ties to set estimation \cite{cuevas2003set} and shape estimation. From a set estimation perspective, estimating a manifold can be seen as estimating the target compact set which is the support of the data generating distribution.

\cite{niyogi_topological_2011} construct an estimator for the support of a data generating distribution that is equivalent to the Devroye-Wise estimator for the support \citep{devroye1980detection}. Although this estimator does not provide guarantees in terms of Hausdorff distance or expected reconstruction error to the true manifold, it is of practical use, and this is mainly the spirit in which we explore the problem in this paper.


\section{Manifold Estimation from Tangent Bundles}

\begin{figure}[ht]
\centering
\includegraphics[trim={1cm 0.5cm 1cm 2cm}, clip, scale=0.5]{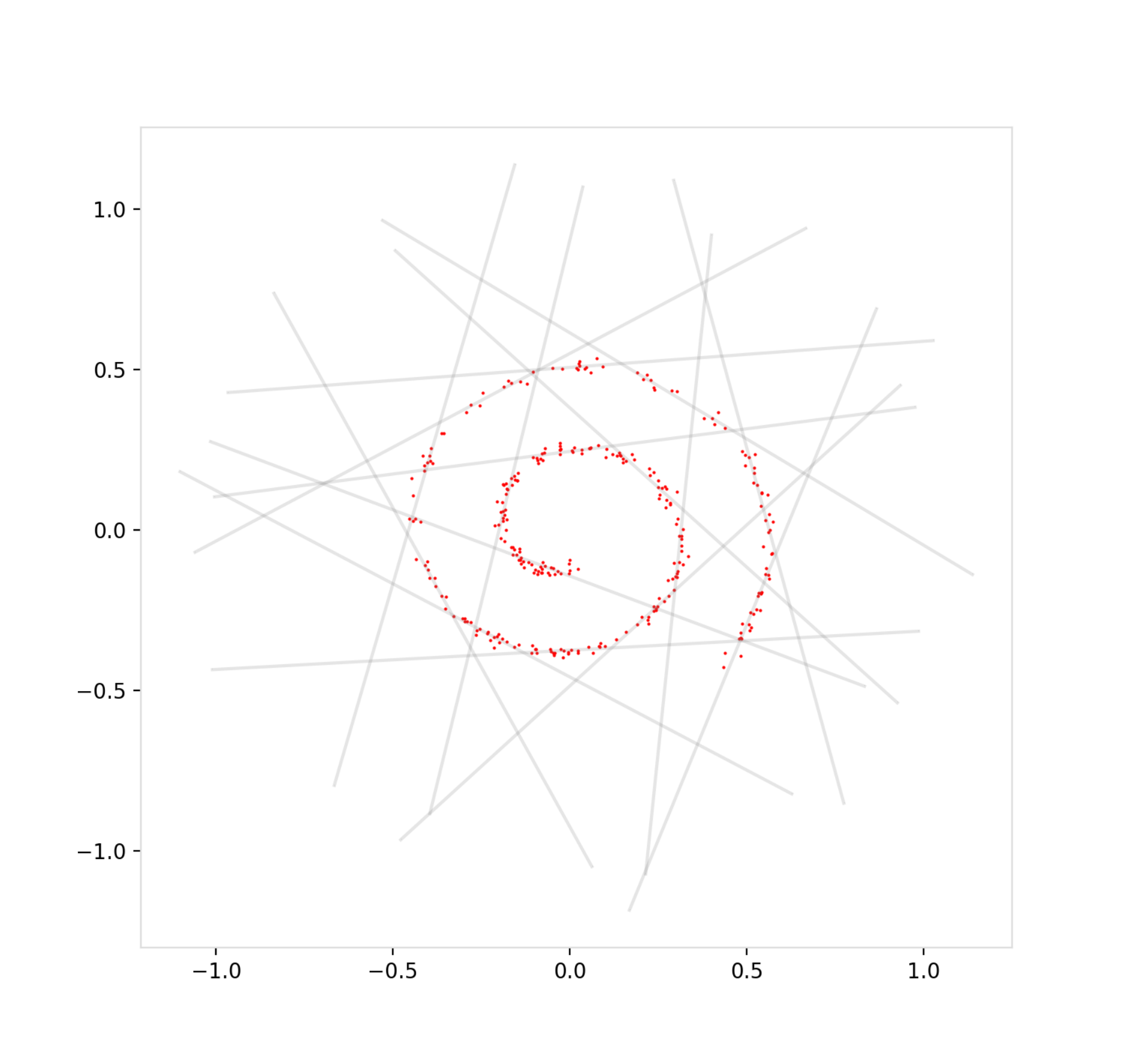}
\caption{``Shards of glass'' estimate from the unbounded tangent spaces of na\"ive local PCA tangent bundle learners.}
\label{fig:naive-lpca}
\end{figure}

The collection of tangent spaces in a tangent bundle give a na\"ive estimate of the manifold. Tangent spaces are, by definition, linear estimates to the manifold surface.  The manifold property of local linearity implies that this estimate is good in some neighborhood.  However, tangent spaces themselves are unbounded, which makes this estimate far too large.  For example, treating the tangent bundle from a TBL na\"ively as a manifold estimator for the spiral and s-curve datasets gives rise to ``shards of glass'' type estimators (see Figure \ref{fig:naive-lpca}) because the tangent spaces simply extend too far.

FNE gives a way of addressing this problem.  The goal of FNE is to consider only the connected subsets ($N_i$) of each tangent space that 1) contain their anchor point and 2) are close to the manifold.  With these subsets, the manifold estimate follows from Equation \ref{eqn:M-hat}.


\subsection{Assignment}

One of the central tools that we require in order to build a manifold estimator from a tangent bundle learner is the assignment relation $A$.  Assignment relations need not be ``hard'' in the sense that they give a unique association between points in $\mathcal{X}$ and $\hat{TB}(\mathcal{X})$, but can be probabilistic or ``soft'' and instead yield a distribution over tangent spaces.  The tangent bundle learners we have discussed and explored here all utilize either hard or soft assignments internally.  The reason is simple: in order to estimate a tangent space, a necessary property is \textit{flatness}.  Only subsets of $\mathcal{X}$ satisfying this property are useful when trying to estimate a tangent space in a certain region.  A relation which yields this is an assignment relation as we have defined it: it associates points in $\mathcal{X}$ with the tangent spaces built from them (and therefore, which the tangent spaces are good linear approximators).  

We can assume two perspectives here in the construction of an assignment relation.  The first is that the tangent bundle learner is a black box: it simply yields $\hat{TB}(\mathcal{X})$ and it is up to us to associate points in $\mathcal{X}$ with elements of $\hat{TB}(\mathcal{X})$.  The risk we run in doing this is that we associate tangent spaces with training points that were not used to estimate them and might be missing essential properties, such as flatness and (in some sense) nearness.  

The second, and we think better, approach is to use the same assignment relation used by the tangent bundle learner itself.  It gives us confidence that the points in $\mathcal{X}$ associated with a tangent space have the necessary properties, such as approximate co-planarity and nearness and therefore can be reliably used for neighborhood estimation in that space.
    
\subsection{Faithful Neighborhood Estimation}
We explore three possible approaches to FNE based on different strategies.  The first two exploit native manifold properties, such as the existence of charts.  The last one is much simpler, but as will be demonstrated in the experiments section, tends not to perform well.

\begin{figure*}[ht]
\vspace{-20pt}
\centering
\subfloat[M1]{\includegraphics[trim={3.5cm 3cm 3cm 4cm}, clip, scale=0.4]{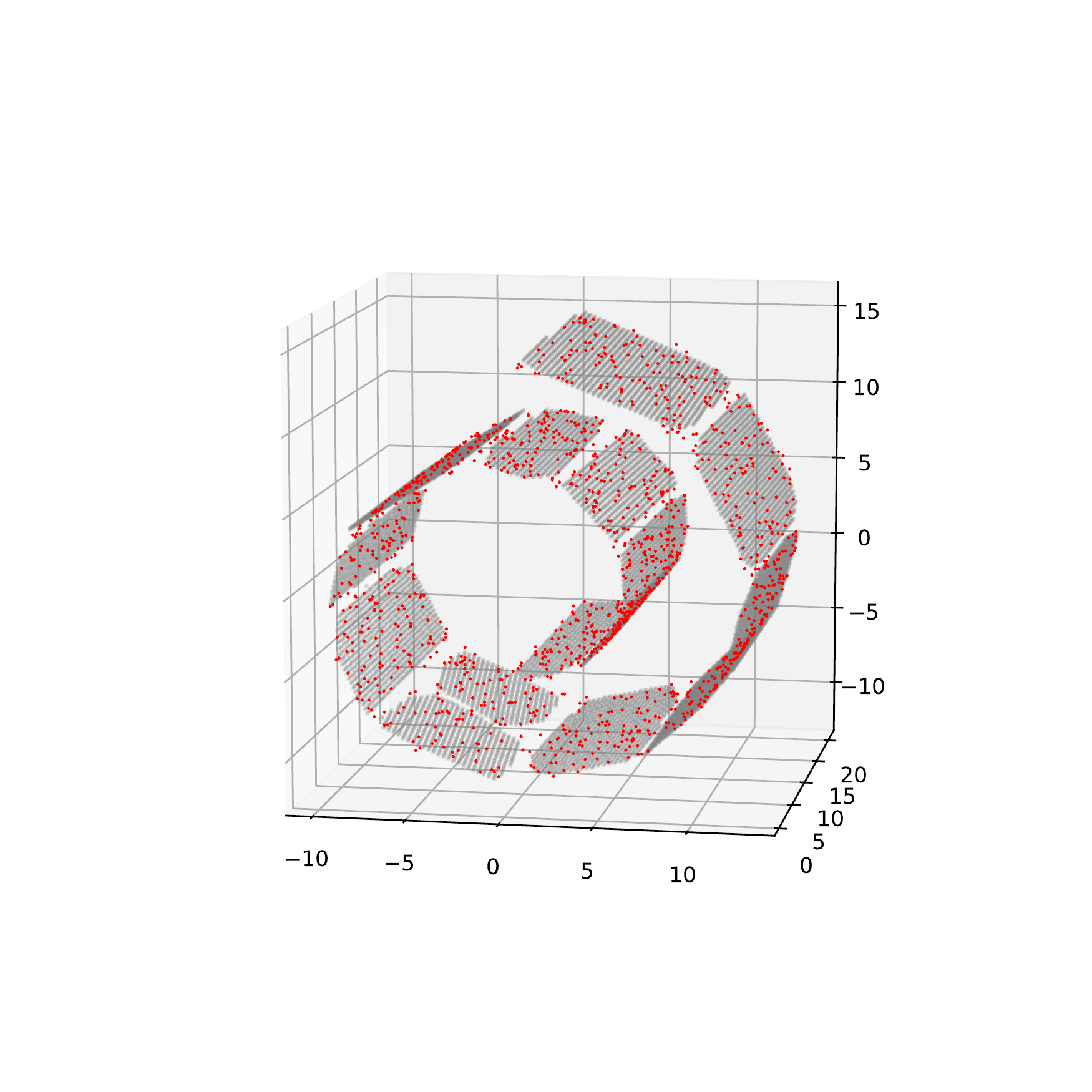}}
\hfill
\subfloat[M2]{\includegraphics[trim={3.5cm 3cm 3cm 4cm}, clip, scale=0.4]{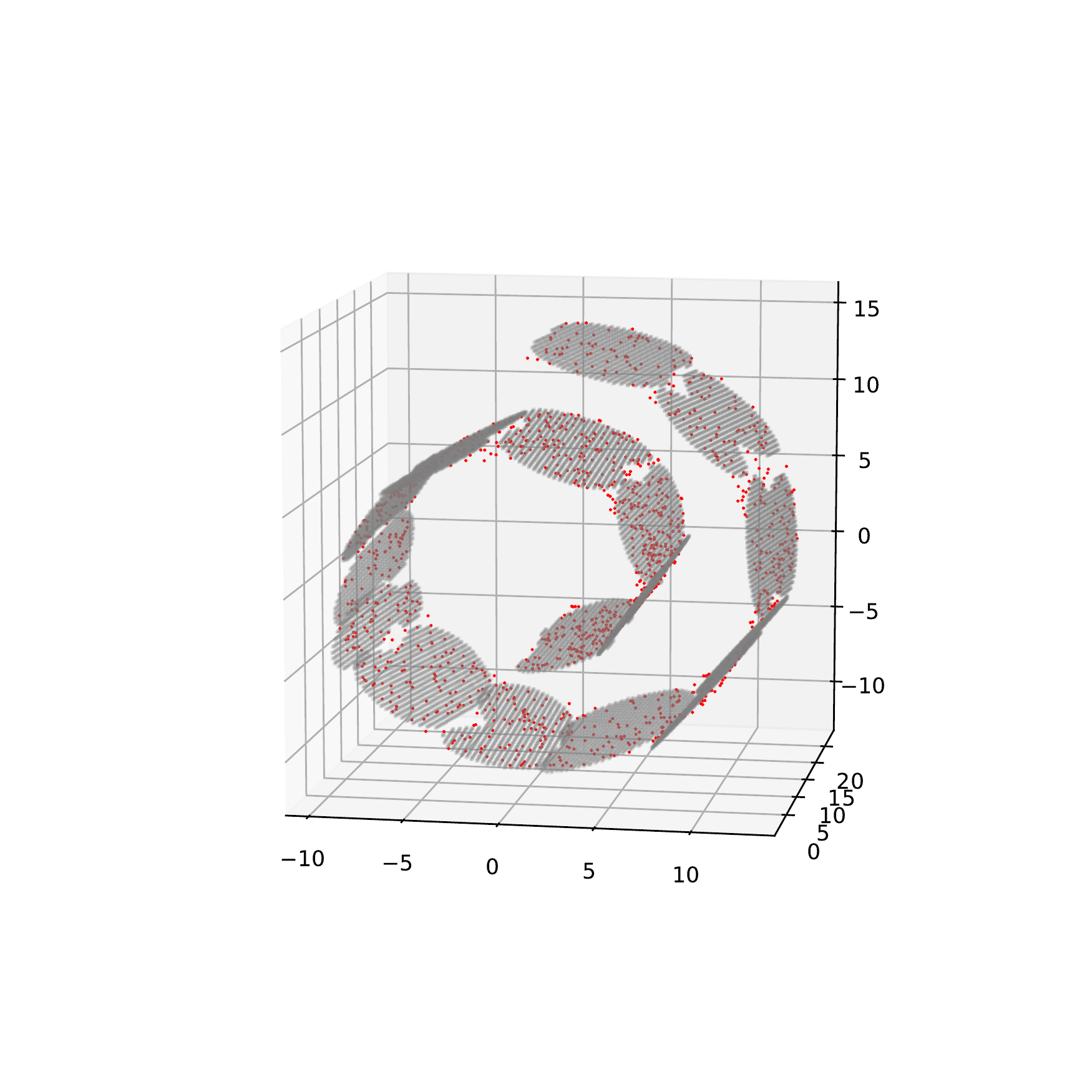}}
\hfill
\subfloat[M3]{\includegraphics[trim={3.5cm 3cm 3cm 4cm}, clip, scale=0.4]{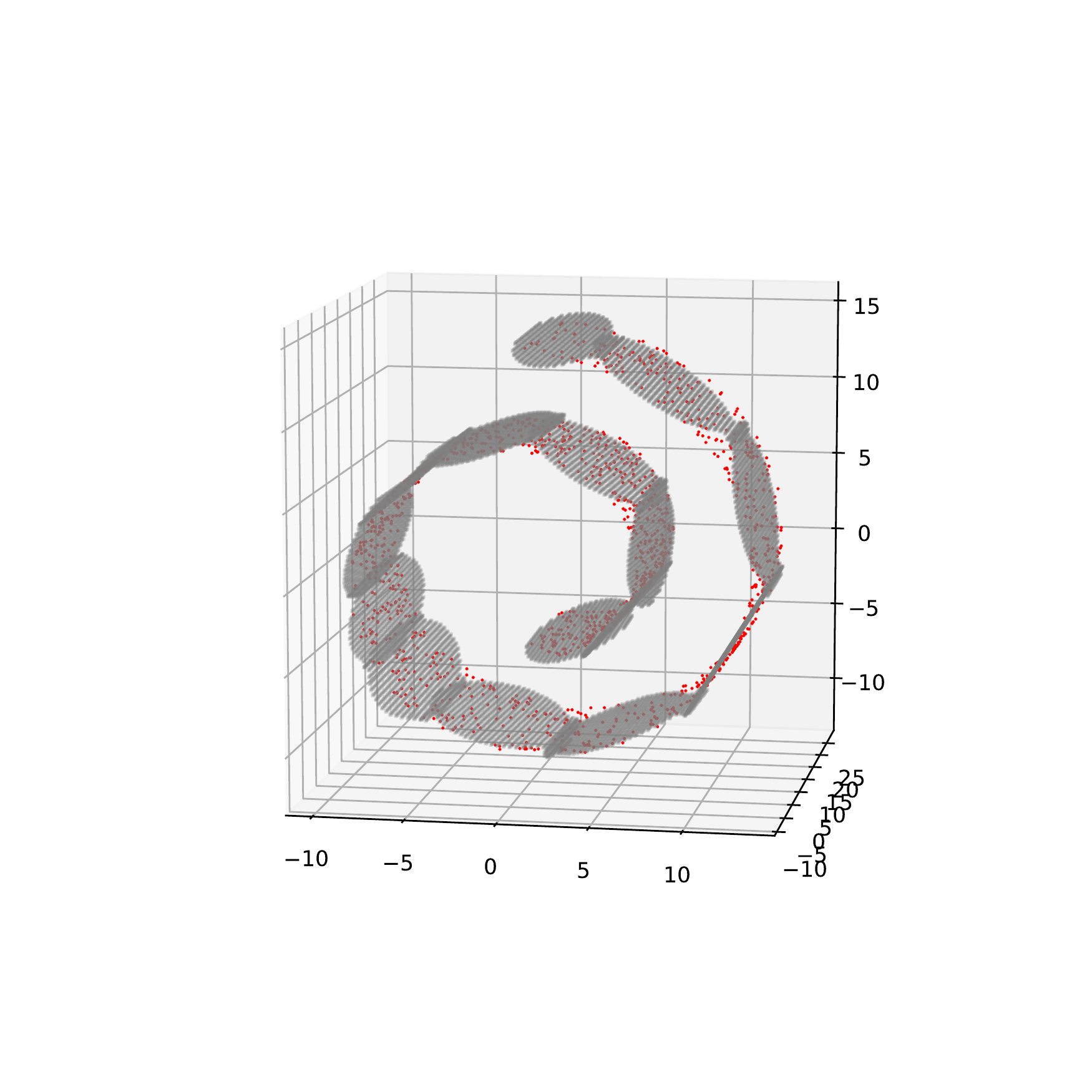}}\\[-1ex]
\caption{Training points in red and the manifold estimates in gray for the swiss roll dataset.}
\label{fig:method_qual}
\vspace{-10pt}
\end{figure*}

Let $(a_i, T_{a_i}) \in \hat{TB}(\mathcal{M})$.  Then, as we have defined it, $A_{T_{a_i}}$ is the set of points in $\mathcal{X}$ for which $T_{a_i}$ is a good linear approximator.  We seek to use this fact to construct a chart  $C_{a_i}: \mathcal{N}_{a_i} \subset \mathcal{M} \rightarrow \mathbb{R}^d $

$$C_{a_i}(x) \coloneqq \mathbf{T}_{a_i}{\mathcal{M}} \cdot (x - a_i)$$
where $\mathbf{T}_{a_i} \in \mathbb{R}^{d \times D}$ is a matrix with rows spanning $T_{a_i}$.  This is the Karhunen-Lo\`eve transform that projects points onto $T_{a_i}$.  The inverse, $C_{a_i}^{-1}: \mathbb{R}^d \rightarrow \mathcal{N}_{a_i}$, then follows as the inverse Karhunen-Lo\`eve transform, defined as 
$$C_{a_i}^{-1}(x) \coloneqq (x'\cdot \mathbf{T}_{a_i}{\mathcal{M}}) + a_i$$
Because of how we have defined the assignment relation $A$, for $x \in A_{T_{a_i}}$, $\|C_{a_i}^{-1}(C_{a_i}(x)) - x\|_2$ should be small.  

The first two methods, M1 and M2, seek to estimate $N_{a_i}$ by the domain of $C_{a_i}$, $\mathcal{N}_{a_i}$.  They each do this by looking at the image of points assigned to the tangent space under the chart for that space,
$$Y_{a_i} \coloneqq C_{a_i}(A_{T_{a_i}})$$
Method 1 (M1) estimates $N_{a_i}$ as the image of the inverse chart of the convex hull of $Y_{a_i}$.  Method 2 (M2) uses $Y_{a_i}$ to build a density estimate in $\mathbb{R}^d$ and extracts a superlevel set from this density using a threshold and estimates $N_{a_i}$ as the image of this superlevel set under the inverse chart, $C_{a_i}^{-1}$.  Method 3, conceptually the simplest, uses a global density estimator in the ambient space and extracts a superlevel set, again, using a threshold.  $N_{a_i}$ is then estimated as the intersection between $T_{a_i}$ and this superlevel set.  Detailed descriptions follow:

\subsubsection{M1: Latent Space Convex Hulls to Faithful Neighborhood Estimates}

Let $H(\cdot)$ be the convex hull of a set of points.  Then, consider $H(Y_{a_i}) \subset \mathbb{R}^d$, the convex hull of points assigned to tangent space $T_{a_i}$ under the chart constructed for that tangent space, $C_{a_i}$.  We estimate $\mathcal{N}_{a_i}$ as 
$$\mathcal{N}_{a_i} = C_{a_i}^{-1}(H(Y_{a_i}))$$ 
and $N_{a_i}$ as $\mathcal{N}_{a_i}$.  See Figure \ref{fig:method_qual}(a) for an example of this method on a canonical datasets.

\subsubsection{M2: Latent Space Density Superlevel Sets to Faithful Neighborhood Estimates} Let $P_{a_i}(\cdot)$ be a probability density defined over $\mathbb{R}^d$ and estimated based on $Y_{a_i}$.  Consider the superlevel set $S_{a_i} \coloneqq P_{a_i}(\mathbb{R}^d) > \alpha_i$, where $\alpha_i$ is some threshold.  Then, we estimate 
$$\mathcal{N}_{a_i} = C_{a_i}^{-1}(S_{a_i})$$ 
and $N_{a_i}$ as $\mathcal{N}_{a_i}$.  See Figure \ref{fig:method_qual}(b) for an example of this method on a canonical dataset using a Gaussian Mixture Model and a tied, global threshold.

\subsubsection{M3: Tangent Space-Ambient Space Density Superlevel Set Intersection for Faithful Neighborhood Estimates}  In method 3, we do not utilize $C_{a_i}$ nor $C_{a_i}^{-1}$.  We assume a global density model $P(X)$ defined over $\mathbb{R}^D$ and estimated based on $\mathcal{X}$.  Then, let $S \coloneqq P(\mathbb{R}^D) > \alpha$ be the superlevel set consisting of the region of $\mathbb{R}^D$ where $P > \alpha$, a global threshold.  We estimate $\mathcal{N}_{a_i}$ as 
$$\mathcal{N}_{a_i} = T_{a_i} \cap S$$ 
and $N_{a_i}$ as $\mathcal{N}_{a_i}$.  See Figure \ref{fig:method_qual}(c) for an example of this method on a canonical dataset.

\section{Experiments}
In this section, we perform experiments to understand the behavior and performance of the proposed manifold estimators. Overall, we report statistics for over a thousand hyperparameter and dataset configurations. For training set size, we used $n \in \{700, 900, 1100, 1300, 1500\}$ for the sine and s-curve datasets and $n \in \{1500, 1700, 1900, 2100, 2300\}$ for the spiral and swiss roll datasets. This differentiation is required as the latter pair represent more complicated manifolds. For the number of components passed to the tangent bundle learner, we vary $n/k \in \{55, 65, 75, 85, 95\}$.

\subsection{Experimental Setup}

\subsubsection{Tangent Bundle Learners}
As a base tangent bundle learner, we chose the Mixtures of Probabilistic Principal Component Analyzers (MoPPCA) method of \cite{tipping_mixtures_1999}, a "mixtures of pancakes" approach.  A simpler approach would be to use k-means clustering followed by PCA in each cluster, but k-means favors spherical clusters and the clusters we seek are flat and close to the tangent space. Another option would be to use the method of \cite{kambhatla1997dimension}, which they refer to as VQPCA.  Like k-means, VQPCA iteratively performs cluster assignment and then cluster centroid calculation.  However, whereas k-means uses Euclidean distance for cluster assignment, favoring spherical clusters, VQPCA uses distance to the tangent space calculated from PCA in each cluster, leading to extremely flat clusters.  However, by \textit{only} considering distance to the tangent space and not, for instance, distance from the centroid, the assignments VQPCA induce can extend far from the centroid and in practice, often cut across the true tangent spaces of the manifold.  MoPPCA  uses a Gaussian Mixture Model (GMM) to perform density estimation in the ambient space. The cluster assignments are based on Mahalanobis distance, a generalization of Euclidean distance that allows each dimension to operate in a different scale.  This allows a compromise between k-means to LPCA \citep{bregler1995nonlinear} and VQPCA \citep{kambhatla1997dimension}: it considers distances in all dimensions, but at different \textit{scales}.  The more a component "pancakes", the more preference distance to the tangent space is given over distance to the centroid.

\subsubsection{Datasets}
We use the 2D spiral, 2D sine wave, 3D s-curve and 3D swiss roll datasets to conduct experiments. 

The 2D spiral dataset was generated from the following distribution of two-dimensional (x, y) points:
\begin{equation}
x = 0.04t \; sin(t) + \epsilon_x,\; y = 0.04t \; cos(t) + \epsilon_y
\end{equation}

where $t \sim U(3, 15)$, $\epsilon_x \sim N(0, 0.01)$, $\epsilon_y \sim N(0, 0.01)$, $U(a,b)$ is uniform in the interval $(a,b)$ and $N(\mu, \sigma)$ is a normal density.

The 2D sine wave dataset was generated from the following distribution of two-dimensional (x, y) points:

\begin{equation}
x = t + \epsilon_x,\; y = sin\left(\frac{2\pi.5.t}{30}\right) + \epsilon_y
\end{equation}

where $t \sim U(3, 15)$, $\epsilon_x \sim N(0, 0.05)$, $\epsilon_y \sim N(0, 0.05)$, $U(a,b)$ and $N(\mu, \sigma)$ are as before.

The 3D s-curve dataset was generated from the following distribution of three-dimensional (x, y, z) points:

\begin{equation}
x = sin(t) + \epsilon_x, \; y = U(0, 2) + \epsilon_y, \; z=sign(t) + (cos(t) - 1) + \epsilon_z
\end{equation}

where $t \sim U(-1.5\pi, 1.5\pi)$, $\epsilon_x \sim N(0, 0.05)$, $\epsilon_y \sim N(0, 0.05)$, $\epsilon_z \sim N(0, 0.05)$, $U(a,b)$ and $N(\mu, \sigma)$ are as before.

The 3D swiss roll dataset was generated from the following distribution of three-dimensional (x, y, z) points:

\begin{equation}
x = t\; cos(t) + \epsilon_x, \; y = U(0, 21) + \epsilon_y, \; z=t \; sin(t) + \epsilon_z
\end{equation}

where $t \sim U(1.5 \pi, 4.5 \pi)$, $\epsilon_x \sim N(0, 0.0005)$, $\epsilon_y \sim N(0, 0.0005)$, $\epsilon_z \sim N(0, 0.0005)$, $U(a,b)$ and $N(\mu, \sigma)$ are as before.

\subsubsection{Evaluation Metrics}
One of the primary questions that arises for manifold estimation is how to evaluate the quality of estimates. What is needed is a distance between sets that penalizes the size of both sets. Expected reconstruction error $\mathcal{\varepsilon}_{\rho}$ has been the focus of some recent work such as \citep{maurer2010k,narayanan2010sample, canas_learning_2012}.
\begin{equation}
\mathcal{\varepsilon}_{\rho}(\hat{\mathcal{M}}) \coloneqq \int_{\mathcal{M}} d\rho(x)\; d^2_{\mathcal{X}}(x, \hat{\mathcal{M}}) 
\end{equation}
Here, $\mathcal{X}$ is a Hilbert space endowed with a Borel probability measure $\rho$ supported over a compact, smooth manifold $\mathcal{M}$ and $d^2_{\mathcal{X}}(x, \hat{\mathcal{M}}) = inf_{x'\in \hat{\mathcal{M}}} d^2_{\mathcal{X}}(x, x')$, with $d_{\mathcal{X}}(x, x') = ||x-x'||$. As \cite{canas_learning_2012} observe, when $\mathcal{\hat{M}} \supset \mathcal{M}$ with $\mathcal{M}$ being the smallest such set with respect to set containment,  the expected error is zero. That is, the error measure does not impose a penalty on the size of $\mathcal{\hat{M}}$. This is problematic because, in the na\"ive formulation, when $N_i$ is the full estimated tangent space, then $\hat{\mathcal{M}}$ is the nonsensical ``shards of glass'' estimate which extends far beyond the manifolds under study here and yet, could, in theory have $\mathcal{\varepsilon}_{\rho}(\mathcal{M},\mathcal{\hat{M}}) = 0$.

Given large uniform random samples from the true manifold and the estimated manifold, $\mathcal{U}$ and $\mathcal{V}$ respectively, a symmetric version of expected reconstruction error would be
\begin{equation}
E(\mathcal{M}, \hat{\mathcal{M}}) = \frac{1}{2} [d^2_{\mathcal{U}}(u, \mathcal{V}) + d^2_{\mathcal{V}}(v, \mathcal{U})]
\end{equation}
where $d(\cdot, \cdot)$ is as above, $u \in \mathcal{U}$ and $v \in \mathcal{V}$. This penalizes the size of both sets. 
Unlike asymmetric Hausdorff, which behaves like $\mathcal{\varepsilon}_{\rho}$, the symmetric Hausdorff distance is another metric that penalizes the size of both sets:
\begin{equation}
H(\mathcal{M},\mathcal{\hat{M}}) = \max(\adjustlimits\sup_{u\in \mathcal{U}} \inf_{v\in \mathcal{V}} \mathrm{d}(u,v), \adjustlimits\sup_{v\in \mathcal{V}} \inf_{u\in \mathcal{U}} \mathrm{d}(u,v))
\end{equation}
For these reasons, we use the symmetric version of expected reconstruction error and Hausdorff distance to measure the quality of manifold estimation in this work. We expect $E$ and $H$ to behave similarly except in cases where there are outliers, to which Hausdorff distance is sensitive.

\subsubsection{Parameters}
M1 does not have any hyperparameters and is straightforward to implement. M2 and M3 have a threshold which determines the superlevel set(s) considered to represent our estimate of the manifold. We optimize over the choice of thresholds on the expected reconstruction error measure using random search over ten iterations in a sensible range on a held out set.
M2 additionally has a hyperparameter in the number of components of the latent space GMMs. A single threshold parameter tied over neighborhoods is optimized over on the expected reconstruction error measure using random search over ten iterations in a sensible range on a held out set.

\begin{figure*}[ht]
\centering
\subfloat[M1]{\includegraphics[trim={0cm 0cm 0cm 0.5cm}, clip, scale=0.4]{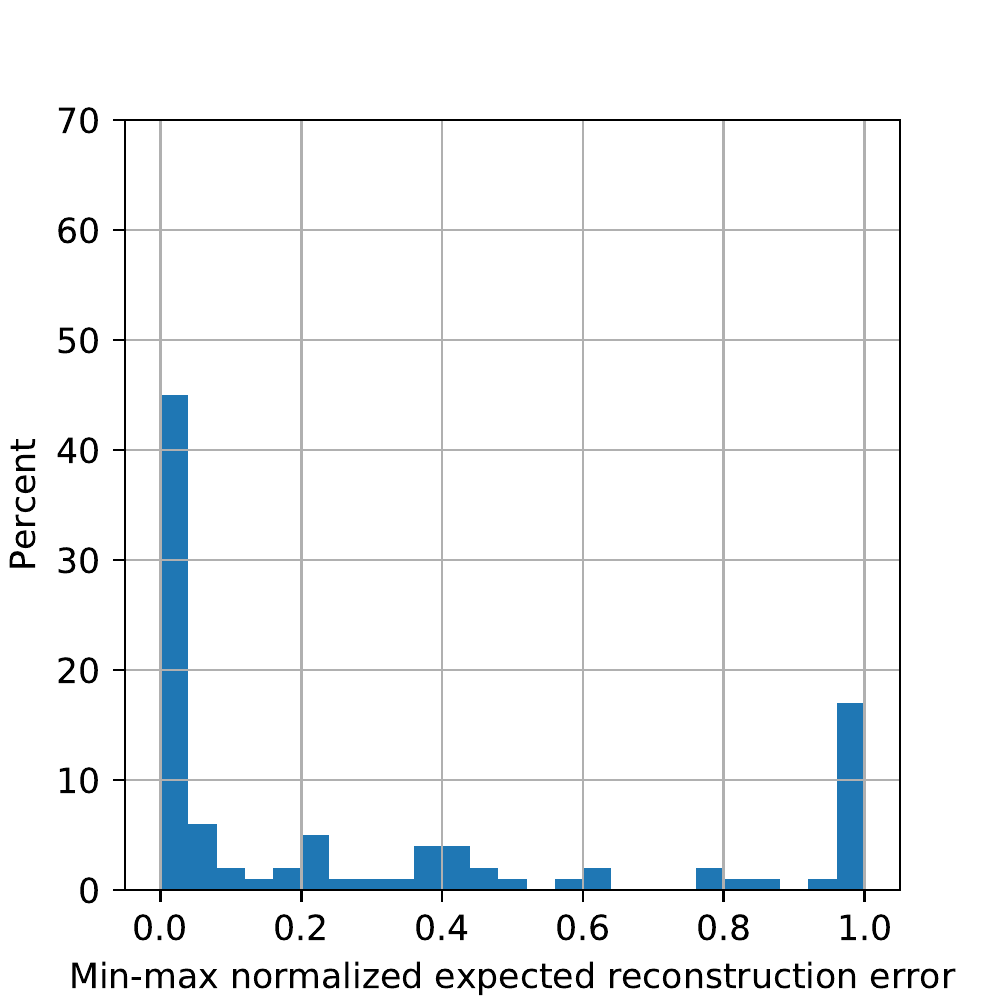}}
\hfill
\subfloat[M2]{\includegraphics[trim={0cm 0cm 0cm 0.5cm}, clip, scale=0.4]{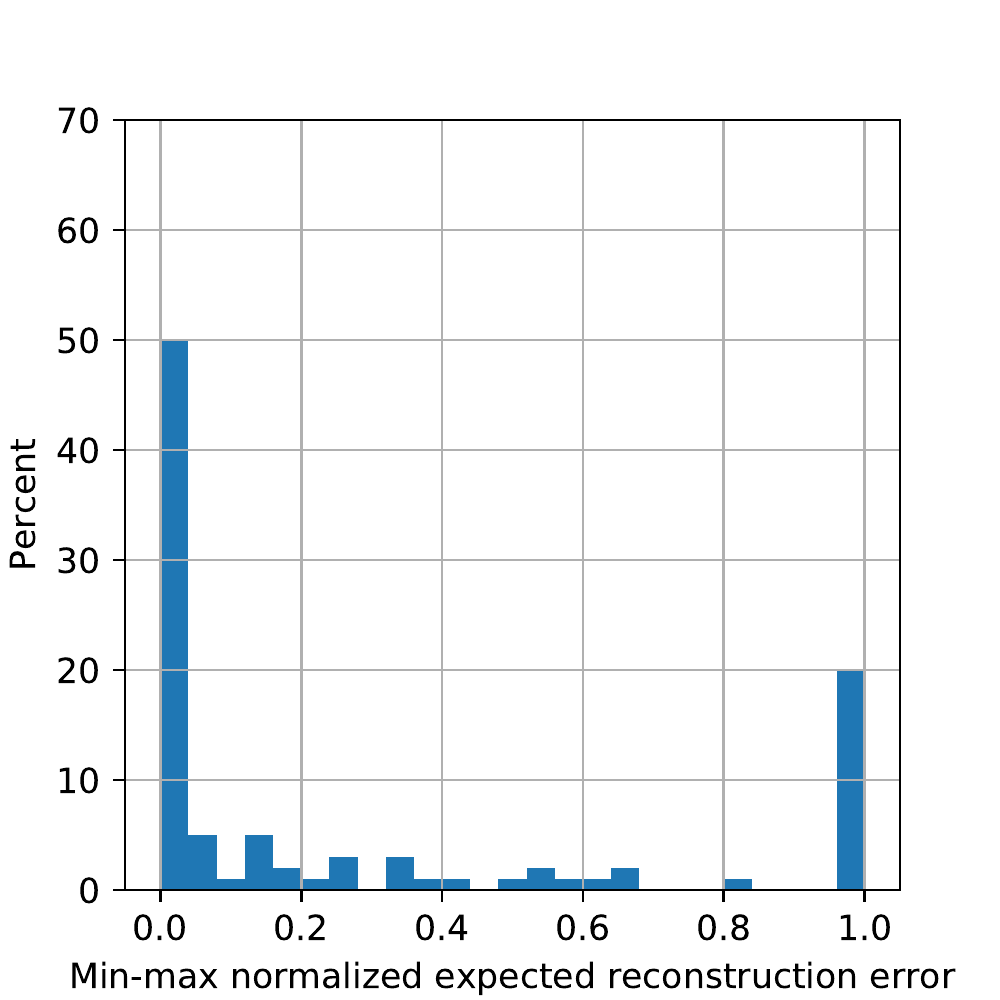}}
\hfill
\subfloat[M3]{\includegraphics[trim={0cm 0cm 0cm 0.5cm}, clip, scale=0.4]{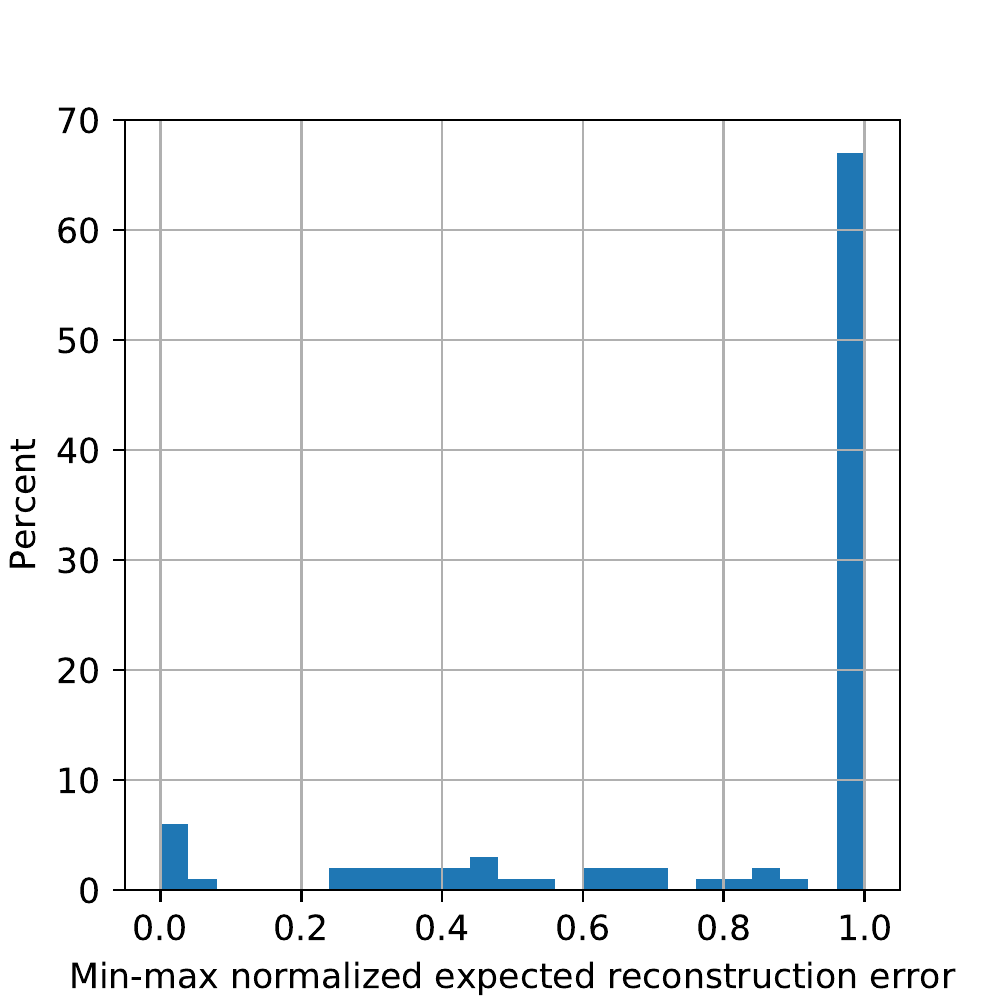}}
\caption{Relative performances of each of the methods over the parameter configurations. A method performs relatively worse as one moves along the x-axis from 0 to 1.}
\label{fig:method-comp}
\end{figure*}

\begin{table}
\centering
\caption{Mean +- 1 standard deviations in expected reconstruction error for the Spiral and Swiss roll datasets, as a function of $n$ and $n/k$.}
\label{tab:ere-table-1}
\resizebox{\linewidth}{!}{%
\begin{tabular}{|c|c|c|c|c|c|c|c|} 
\hline
\multicolumn{2}{|c|}{Dataset} & \multicolumn{3}{c|}{Spiral}                                                       & \multicolumn{3}{c|}{Swiss roll}                                                    \\ 
\hline
n/k                 & n       & M1                           & M2                           & M3                  & M1                           & M2                           & M3                   \\ 
\hline
\multirow{5}{*}{55} & 1500    & \textbf{6.86E-5 $\pm$ 2E-6}  & 6.88E-5 $\pm$ 2E-6           & 8.25E-5 $\pm$ 2E-5  & 3.71E-1 $\pm$ 9E-3           & \textbf{3.46E-1 $\pm$ 1E-2}  & 3.77E-1 $\pm$ 2E-2   \\ 
\cline{2-8}
                    & 1700    & 6.56E-5 $\pm$ 2E-6           & \textbf{6.53E-5 $\pm$ 3E-6}  & 7.40E-5 $\pm$ 1E-5  & 3.53E-1 $\pm$ 1E-2           & \textbf{3.42E-1 $\pm$ 5E-3}  & 3.65E-1 $\pm$ 1E-2   \\ 
\cline{2-8}
                    & 1900    & \textbf{6.31E-5 $\pm$ 2E-6}  & 6.45E-5 $\pm$ 2E-6           & 6.95E-5 $\pm$ 3E-6  & 3.50E-1 $\pm$ 1E-2           & \textbf{3.35E-1 $\pm$ 8E-3}  & 3.65E-1 $\pm$ 2E-2   \\ 
\cline{2-8}
                    & 2100    & \textbf{6.20E-5 $\pm$ 2E-6}  & 6.35E-5 $\pm$ 2E-6           & 7.55E-5 $\pm$ 2E-5  & 3.55E-1 $\pm$ 1E-2           & \textbf{3.34E-1 $\pm$ 8E-3}  & 3.60E-1 $\pm$ 2E-2   \\ 
\cline{2-8}
                    & 2300    & \textbf{6.12E-5 $\pm$ 1E-6}  & 6.21E-5 $\pm$ 2E-6           & 7.09E-5 $\pm$ 1E-5  & 3.47E-1 $\pm$ 9E-3           & \textbf{3.34E-1 $\pm$ 9E-3}  & 3.51E-1 $\pm$ 1E-2   \\ 
\hline
\multirow{5}{*}{65} & 1500    & 7.52E-5 $\pm$ 3E-6           & \textbf{7.44E-5 $\pm$ 3E-6}  & 9.13E-5 $\pm$ 1E-5  & \textbf{3.54E-1 $\pm$ 1E-2}  & 3.62E-1 $\pm$ 2E-2           & 3.81E-1 $\pm$ 2E-2   \\ 
\cline{2-8}
                    & 1700    & \textbf{6.79E-5 $\pm$ 2E-6}  & 6.82E-5 $\pm$ 3E-6           & 7.65E-5 $\pm$ 8E-6  & \textbf{3.52E-1 $\pm$ 8E-3}  & 3.53E-1 $\pm$ 2E-2           & 3.95E-1 $\pm$ 4E-2   \\ 
\cline{2-8}
                    & 1900    & \textbf{6.43E-5 $\pm$ 1E-6}  & 6.64E-5 $\pm$ 2E-6           & 6.96E-5 $\pm$ 3E-6  & 3.49E-1 $\pm$ 1E-2           & \textbf{3.42E-1 $\pm$ 1E-2}  & 3.67E-1 $\pm$ 2E-2   \\ 
\cline{2-8}
                    & 2100    & 6.42E-5 $\pm$ 3E-6           & \textbf{6.35E-5 $\pm$ 1E-6}  & 6.93E-5 $\pm$ 4E-6  & 3.54E-1 $\pm$ 1E-2           & \textbf{3.35E-1 $\pm$ 1E-2}  & 3.59E-1 $\pm$ 2E-2   \\ 
\cline{2-8}
                    & 2300    & \textbf{6.10E-5 $\pm$ 2E-6}  & 6.32E-5 $\pm$ 2E-6           & 6.94E-5 $\pm$ 8E-6  & 3.47E-1 $\pm$ 1E-2           & \textbf{3.28E-1 $\pm$ 9E-3}  & 3.57E-1 $\pm$ 3E-2   \\ 
\hline
\multirow{5}{*}{75} & 1500    & \textbf{8.02E-5 $\pm$ 3E-6}  & 8.21E-5 $\pm$ 6E-6           & 9.75E-5 $\pm$ 1E-5  & \textbf{3.60E-1 $\pm$ 1E-2}  & 3.92E-1 $\pm$ 8E-2           & 4.11E-1 $\pm$ 3E-2   \\ 
\cline{2-8}
                    & 1700    & \textbf{7.46E-5 $\pm$ 2E-6}  & 7.53E-5 $\pm$ 3E-6           & 8.50E-5 $\pm$ 8E-6  & 3.55E-1 $\pm$ 1E-2           & \textbf{3.50E-1 $\pm$ 6E-3}  & 3.90E-1 $\pm$ 6E-2   \\ 
\cline{2-8}
                    & 1900    & 6.99E-5 $\pm$ 3E-6           & \textbf{6.88E-5 $\pm$ 3E-6}  & 7.82E-5 $\pm$ 8E-6  & 3.50E-1 $\pm$ 1E-2           & \textbf{3.49E-1 $\pm$ 2E-2}  & 3.92E-1 $\pm$ 4E-2   \\ 
\cline{2-8}
                    & 2100    & \textbf{6.48E-5 $\pm$ 3E-6}  & 6.59E-5 $\pm$ 2E-6           & 7.54E-5 $\pm$ 7E-6  & 3.50E-1 $\pm$ 1E-2           & \textbf{3.38E-1 $\pm$ 1E-2}  & 3.63E-1 $\pm$ 2E-2   \\ 
\cline{2-8}
                    & 2300    & \textbf{6.40E-5 $\pm$ 2E-6}  & 6.47E-5 $\pm$ 2E-6           & 6.72E-5 $\pm$ 3E-6  & \textbf{3.38E-1 $\pm$ 1E-2}  & 3.46E-1 $\pm$ 2E-2           & 3.83E-1 $\pm$ 5E-2   \\ 
\hline
\multirow{5}{*}{85} & 1500    & \textbf{9.55E-5 $\pm$ 6E-6}  & 1.01E-4 $\pm$ 8E-6           & 1.34E-4 $\pm$ 3E-5  & 4.33E-1 $\pm$ 1E-1           & \textbf{4.22E-1 $\pm$ 1E-1}  & 4.24E-1 $\pm$ 5E-2   \\ 
\cline{2-8}
                    & 1700    & 8.10E-5 $\pm$ 5E-6           & \textbf{7.89E-5 $\pm$ 4E-6}  & 1.06E-4 $\pm$ 2E-5  & 3.81E-1 $\pm$ 9E-2           & \textbf{3.65E-1 $\pm$ 1E-2}  & 3.94E-1 $\pm$ 3E-2   \\ 
\cline{2-8}
                    & 1900    & \textbf{7.37E-5 $\pm$ 3E-6}  & 7.43E-5 $\pm$ 4E-6           & 8.76E-5 $\pm$ 1E-5  & 3.50E-1 $\pm$ 9E-3           & \textbf{3.49E-1 $\pm$ 1E-2}  & 3.91E-1 $\pm$ 5E-2   \\ 
\cline{2-8}
                    & 2100    & 7.13E-5 $\pm$ 2E-6           & \textbf{6.94E-5 $\pm$ 1E-6}  & 7.71E-5 $\pm$ 4E-6  & 3.49E-1 $\pm$ 9E-3           & \textbf{3.49E-1 $\pm$ 8E-3}  & 3.71E-1 $\pm$ 2E-2   \\ 
\cline{2-8}
                    & 2300    & 6.72E-5 $\pm$ 2E-6           & \textbf{6.62E-5 $\pm$ 2E-6}  & 6.92E-5 $\pm$ 4E-6  & 3.44E-1 $\pm$ 1E-2           & \textbf{3.33E-1 $\pm$ 5E-3}  & 3.65E-1 $\pm$ 1E-2   \\ 
\hline
\multirow{5}{*}{95} & 1500    & \textbf{1.13E-4 $\pm$ 1E-5}  & 1.21E-4 $\pm$ 1E-5           & 1.52E-4 $\pm$ 2E-5  & 5.45E-1 $\pm$ 2E-1           & \textbf{4.74E-1 $\pm$ 1E-1}  & 4.91E-1 $\pm$ 1E-1   \\ 
\cline{2-8}
                    & 1700    & \textbf{8.98E-5 $\pm$ 4E-6}  & 9.62E-5 $\pm$ 7E-6           & 1.33E-4 $\pm$ 4E-5  & 4.97E-1 $\pm$ 2E-1           & \textbf{4.33E-1 $\pm$ 1E-1}  & 4.90E-1 $\pm$ 1E-1   \\ 
\cline{2-8}
                    & 1900    & 7.97E-5 $\pm$ 5E-6           & \textbf{7.75E-5 $\pm$ 3E-6}  & 9.70E-5 $\pm$ 1E-5  & 3.59E-1 $\pm$ 9E-3           & \textbf{3.55E-1 $\pm$ 1E-2}  & 3.92E-1 $\pm$ 3E-2   \\ 
\cline{2-8}
                    & 2100    & 7.58E-5 $\pm$ 4E-6           & \textbf{7.16E-5 $\pm$ 2E-6}  & 9.23E-5 $\pm$ 1E-5  & 3.48E-1 $\pm$ 9E-3           & \textbf{3.46E-1 $\pm$ 9E-3}  & 3.80E-1 $\pm$ 2E-2   \\ 
\cline{2-8}
                    & 2300    & 7.03E-5 $\pm$ 2E-6           & \textbf{6.87E-5 $\pm$ 1E-6}  & 7.50E-5 $\pm$ 5E-6  & \textbf{3.43E-1 $\pm$ 6E-3}  & 3.49E-1 $\pm$ 1E-2           & 4.00E-1 $\pm$ 3E-2   \\
\hline
\end{tabular}
}
\end{table}

\begin{table}
\centering
\caption{Mean +- 1 standard deviations in expected reconstruction error for the Sine wave and S-curve datasets, as a function of $n$ and $n/k$.}
\label{tab:ere-table-2}
\resizebox{\linewidth}{!}{%
\begin{tabular}{|c|c|c|c|c|c|c|c|} 
\hline
\multicolumn{2}{|c|}{Dataset}   & \multicolumn{3}{c|}{Sine}                                                         & \multicolumn{3}{c|}{S-curve}                                                                \\ 
\hline
n/k                          & n    & M1                           & M2                           & M3                  & M1                           & M2                           & M3                            \\ 
\hline
\multirow{5}{*}{55}          & 700  & \textbf{1.69E-3 $\pm$ 5E-5}  & 1.73E-3 $\pm$ 9E-5           & 1.98E-3 $\pm$ 1E-4  & \textbf{7.25E-3 $\pm$ 5E-4}  & 7.29E-3 $\pm$ 7E-4           & 7.33E-3 $\pm$ 3E-4            \\ 
\cline{2-8}
                             & 900  & \textbf{1.51E-3 $\pm$ 8E-5}  & 1.57E-3 $\pm$ 6E-5           & 1.70E-3 $\pm$ 2E-4  & 6.64E-3 $\pm$ 2E-4           & 6.53E-3 $\pm$ 3E-4           & \textbf{6.49E-3 $\pm$ 2E-4}   \\ 
\cline{2-8}
                             & 1100 & \textbf{1.41E-3 $\pm$ 5E-5}  & 1.48E-3 $\pm$ 4E-5           & 1.56E-3 $\pm$ 9E-5  & 6.41E-3 $\pm$ 2E-4           & \textbf{6.26E-3 $\pm$ 3E-4}  & 6.32E-3 $\pm$ 1E-4            \\ 
\cline{2-8}
                             & 1300 & \textbf{1.41E-3 $\pm$ 6E-5}  & 1.47E-3 $\pm$ 6E-5           & 1.49E-3 $\pm$ 8E-5  & 6.42E-3 $\pm$ 2E-4           & \textbf{5.97E-3 $\pm$ 1E-4}  & 6.15E-3 $\pm$ 1E-4            \\ 
\cline{2-8}
                             & 1500 & \textbf{1.35E-3 $\pm$ 3E-5}  & 1.47E-3 $\pm$ 6E-5           & 1.49E-3 $\pm$ 8E-5  & 6.14E-3 $\pm$ 8E-5           & \textbf{6.01E-3 $\pm$ 8E-5}  & 6.09E-3 $\pm$ 1E-4            \\ 
\hline
\multirow{5}{*}{65}          & 700  & \textbf{1.90E-3 $\pm$ 1E-4}  & 2.23E-3 $\pm$ 3E-4           & 2.50E-3 $\pm$ 4E-4  & 7.66E-3 $\pm$ 5E-4           & \textbf{7.49E-3 $\pm$ 9E-4}  & 7.99E-3 $\pm$ 6E-4            \\ 
\cline{2-8}
                             & 900  & \textbf{1.58E-3 $\pm$ 7E-5}  & 1.65E-3 $\pm$ 8E-5           & 1.86E-3 $\pm$ 1E-4  & 6.88E-3 $\pm$ 5E-4           & 6.63E-3 $\pm$ 5E-4           & \textbf{6.62E-3 $\pm$ 2E-4}   \\ 
\cline{2-8}
                             & 1100 & \textbf{1.47E-3 $\pm$ 8E-5}  & 1.53E-3 $\pm$ 6E-5           & 1.63E-3 $\pm$ 1E-4  & 6.49E-3 $\pm$ 2E-4           & \textbf{6.30E-3 $\pm$ 3E-4}  & 6.41E-3 $\pm$ 2E-4            \\ 
\cline{2-8}
                             & 1300 & 1.43E-3 $\pm$ 4E-5           & \textbf{1.42E-3 $\pm$ 6E-5}  & 1.50E-3 $\pm$ 6E-5  & 6.36E-3 $\pm$ 2E-4           & \textbf{6.13E-3 $\pm$ 2E-4}  & 6.27E-3 $\pm$ 2E-4            \\ 
\cline{2-8}
                             & 1500 & \textbf{1.41E-3 $\pm$ 6E-5}  & 1.44E-3 $\pm$ 5E-5           & 1.54E-3 $\pm$ 8E-5  & 6.20E-3 $\pm$ 1E-4           & \textbf{5.97E-3 $\pm$ 1E-4}  & 6.20E-3 $\pm$ 2E-4            \\ 
\hline
\multirow{5}{*}{75}          & 700  & \textbf{2.07E-3 $\pm$ 2E-4}  & 2.40E-3 $\pm$ 4E-4           & 2.76E-3 $\pm$ 3E-4  & 8.09E-3 $\pm$ 1E-3           & \textbf{7.94E-3 $\pm$ 6E-4}  & 8.16E-3 $\pm$ 1E-3            \\ 
\cline{2-8}
                             & 900  & \textbf{1.68E-3 $\pm$ 7E-5}  & 1.70E-3 $\pm$ 9E-5           & 1.81E-3 $\pm$ 1E-4  & \textbf{7.00E-3 $\pm$ 4E-4}  & 7.07E-3 $\pm$ 4E-4           & 7.23E-3 $\pm$ 8E-4            \\ 
\cline{2-8}
                             & 1100 & \textbf{1.53E-3 $\pm$ 6E-5}  & 1.57E-3 $\pm$ 7E-5           & 1.69E-3 $\pm$ 2E-4  & 6.54E-3 $\pm$ 3E-4           & \textbf{6.54E-3 $\pm$ 3E-4}  & 6.80E-3 $\pm$ 6E-4            \\ 
\cline{2-8}
                             & 1300 & \textbf{1.46E-3 $\pm$ 8E-5}  & 1.49E-3 $\pm$ 6E-5           & 1.52E-3 $\pm$ 8E-5  & 6.32E-3 $\pm$ 2E-4           & \textbf{6.12E-3 $\pm$ 2E-4}  & 6.23E-3 $\pm$ 2E-4            \\ 
\cline{2-8}
                             & 1500 & \textbf{1.39E-3 $\pm$ 5E-5}  & 1.43E-3 $\pm$ 5E-5           & 1.45E-3 $\pm$ 6E-5  & 6.22E-3 $\pm$ 2E-4           & \textbf{6.01E-3 $\pm$ 2E-4}  & 6.15E-3 $\pm$ 2E-4            \\ 
\hline
\multirow{5}{*}{85}          & 700  & \textbf{2.61E-3 $\pm$ 2E-4}  & 2.96E-3 $\pm$ 3E-4           & 3.53E-3 $\pm$ 6E-4  & 9.06E-3 $\pm$ 1E-3           & \textbf{8.46E-3 $\pm$ 1E-3}  & 8.93E-3 $\pm$ 6E-4            \\ 
\cline{2-8}
                             & 900  & \textbf{1.81E-3 $\pm$ 2E-4}  & 1.97E-3 $\pm$ 1E-4           & 2.35E-3 $\pm$ 4E-4  & \textbf{7.23E-3 $\pm$ 4E-4}  & 7.69E-3 $\pm$ 1E-3           & 7.65E-3 $\pm$ 6E-4            \\ 
\cline{2-8}
                             & 1100 & \textbf{1.64E-3 $\pm$ 5E-5}  & 1.70E-3 $\pm$ 1E-4           & 1.90E-3 $\pm$ 3E-4  & \textbf{6.48E-3 $\pm$ 3E-4}  & 6.55E-3 $\pm$ 4E-4           & 6.82E-3 $\pm$ 4E-4            \\ 
\cline{2-8}
                             & 1300 & \textbf{1.49E-3 $\pm$ 6E-5}  & 1.50E-3 $\pm$ 1E-4           & 1.69E-3 $\pm$ 1E-4  & 6.29E-3 $\pm$ 2E-4           & \textbf{6.14E-3 $\pm$ 2E-4}  & 6.36E-3 $\pm$ 2E-4            \\ 
\cline{2-8}
                             & 1500 & \textbf{1.41E-3 $\pm$ 4E-5}  & 1.45E-3 $\pm$ 4E-5           & 1.53E-3 $\pm$ 1E-4  & \textbf{6.15E-3 $\pm$ 1E-4}  & 6.21E-3 $\pm$ 3E-4           & 6.36E-3 $\pm$ 4E-4            \\ 
\hline
\multirow{5}{*}{95}          & 700  & \textbf{2.89E-3 $\pm$ 1E-4}  & 3.40E-3 $\pm$ 5E-4           & 4.45E-3 $\pm$ 1E-3  & \textbf{1.01E-2 $\pm$ 1E-3}  & 1.09E-2 $\pm$ 1E-3           & 1.04E-2 $\pm$ 2E-3            \\ 
\cline{2-8}
                             & 900  & \textbf{2.00E-3 $\pm$ 2E-4}  & 2.27E-3 $\pm$ 3E-4           & 2.67E-3 $\pm$ 3E-4  & \textbf{7.74E-3 $\pm$ 7E-4}  & 8.14E-3 $\pm$ 5E-4           & 8.26E-3 $\pm$ 8E-4            \\ 
\cline{2-8}
                             & 1100 & 1.78E-3 $\pm$ 1E-4           & \textbf{1.73E-3 $\pm$ 3E-5}  & 1.95E-3 $\pm$ 2E-4  & 6.98E-3 $\pm$ 7E-4           & \textbf{6.88E-3 $\pm$ 7E-4}  & 7.23E-3 $\pm$ 6E-4            \\ 
\cline{2-8}
                             & 1300 & \textbf{1.52E-3 $\pm$ 4E-5}  & 1.61E-3 $\pm$ 4E-5           & 1.65E-3 $\pm$ 8E-5  & \textbf{6.33E-3 $\pm$ 2E-4}  & 6.54E-3 $\pm$ 5E-4           & 6.50E-3 $\pm$ 3E-4            \\ 
\cline{2-8}
                             & 1500 & 1.48E-3 $\pm$ 5E-5           & \textbf{1.47E-3 $\pm$ 3E-5}  & 1.56E-3 $\pm$ 1E-4  & 6.24E-3 $\pm$ 2E-4           & 6.34E-3 $\pm$ 3E-4           & \textbf{6.21E-3 $\pm$ 3E-4}   \\
\hline
\end{tabular}
}
\end{table}

We ran experiments to visualize the estimated manifold for each of the methods on each of the datasets. Figure \ref{fig:method_qual} shows the best manifold estimates in terms of expected reconstruction error over ten repetitions of optimizing over hyperparameters using methods M1, M2 and M3 respectively on the spiral dataset. See supplementary material for other datasets. The results shown correspond to the least number of components ($n/k=95$ and $n=1500$ for the spiral and swiss roll datasets and $n=700$ for the sine wave and s-curve) so as to enable the reader to get a sense of how each method works. 

\subsection{Faithful Neighborhood Estimation Method Evaluation}

To compare the relative performance of the methods, we did the following. For each configuration of dataset, $n/k$ and $n$, we picked the best performing expected reconstruction error over 10 repetitions, optimizing over the hyperparameters as before. Then, across methods, we min-max normalized the expected reconstruction errors to study their relative performances. Figure \ref{fig:method-comp} shows the distribution over the normalized reconstruction errors for each of the three methods. Some critical observations can be made. Firstly, M1 performs close to the best in over 40\% of dataset configurations across datasets and M2 is the overall winner, performing best in 50\% of configurations. Although M2 performs close to the worst of the three for more of the configurations than M1, the distribution for M2 is skewed to the left, making it better overall. Secondly, M3, being the least principled, performs close to the worst in almost 70\% of dataset configurations and is the best performing of the three only a nominal percent of the time. 
Thirdly, as can be seen in Tables \ref{tab:ere-table-1} and \ref{tab:ere-table-2}, M1 tends to perform better on the 2D datasets when compared to the 3D datasets. Lastly, M2 is the best performing even though density threshold to determine the superlevel sets was treated as a tied hyperparameter over the tangent spaces. 

\begin{figure*}[ht]
\vspace{-15pt}
\centering
\subfloat[The spiral.]{\includegraphics[trim={4cm 2cm 2cm 4cm}, clip, scale=0.4]{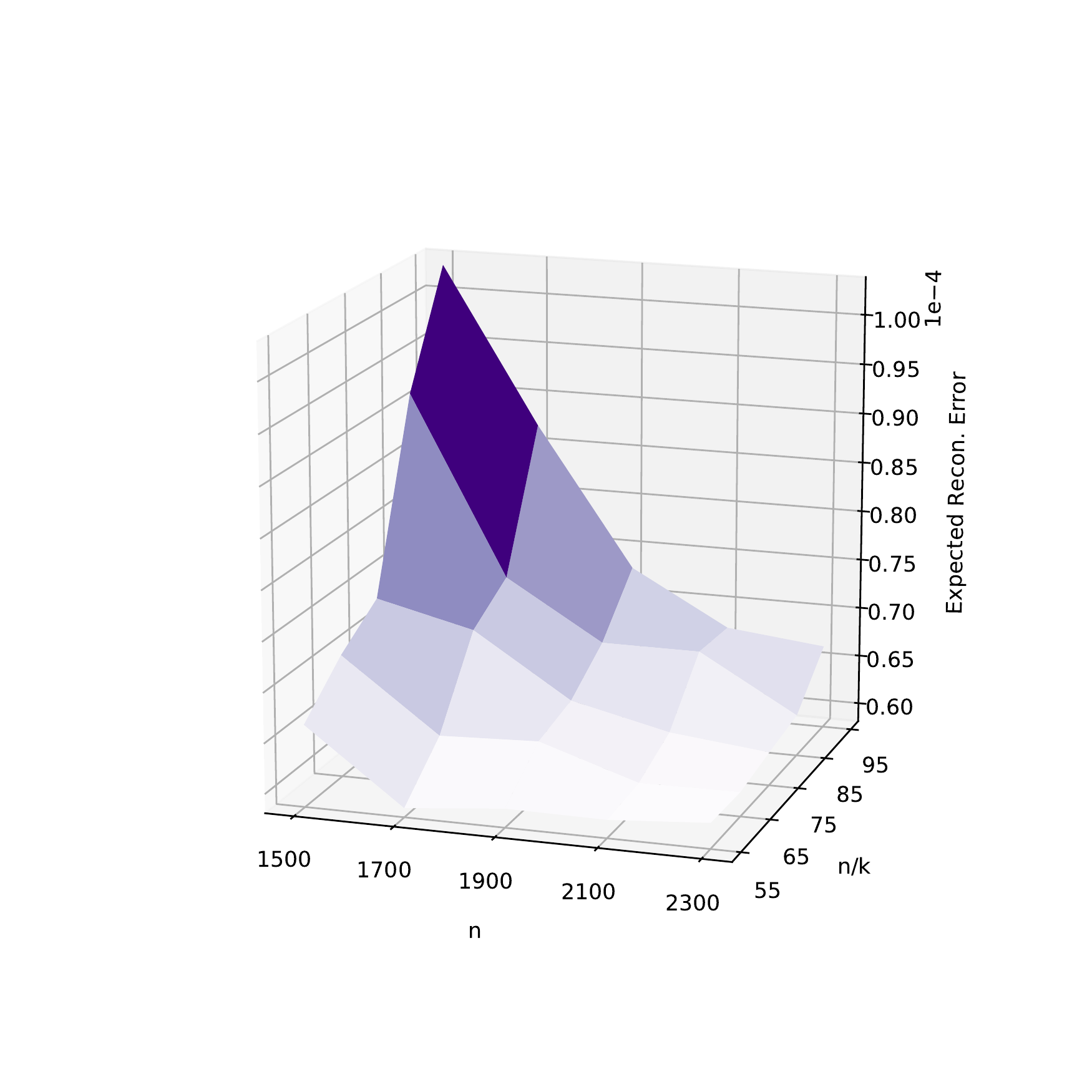}}
\hfill
\subfloat[The sine wave.]{\includegraphics[trim={4cm 2cm 2cm 4cm}, clip, scale=0.4]{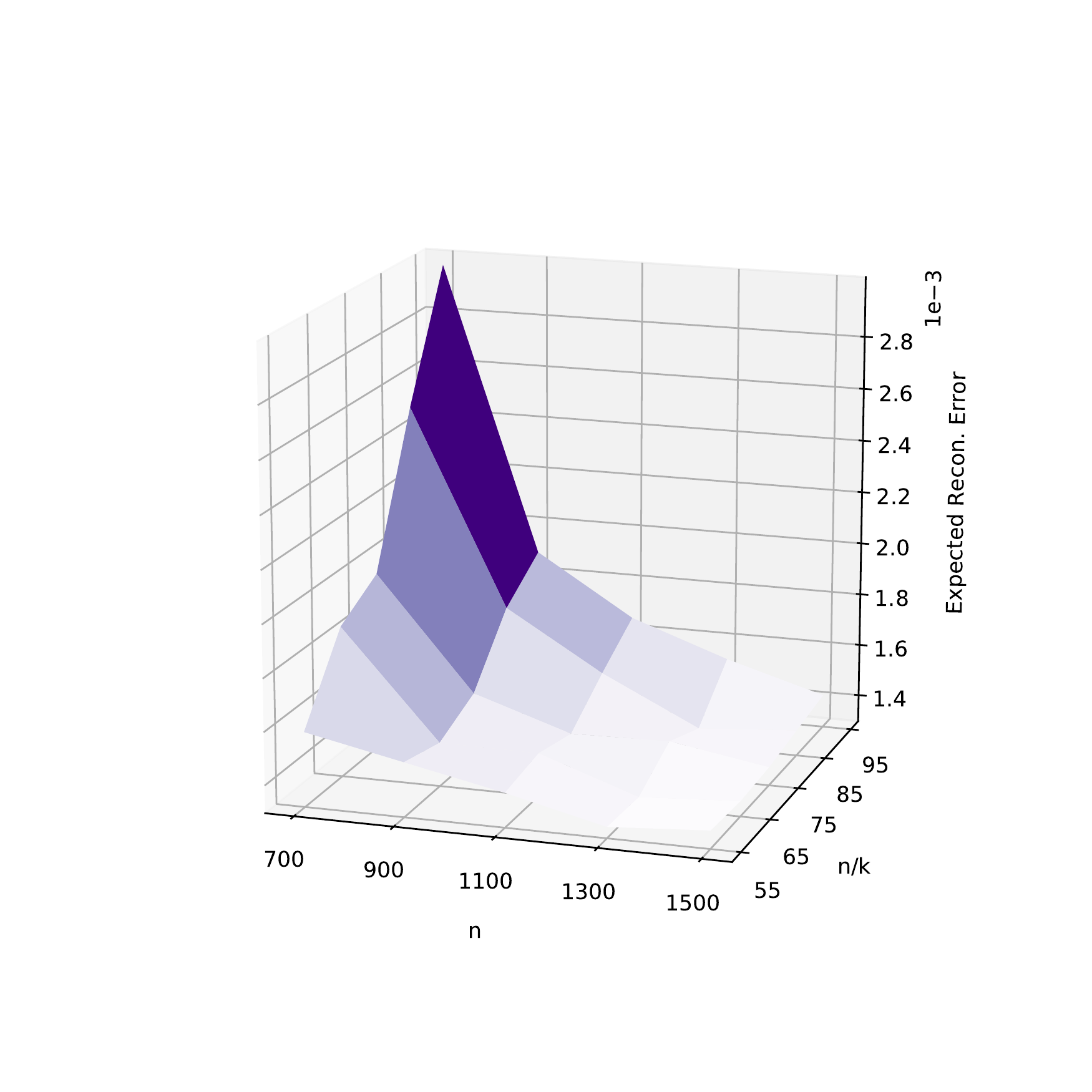}}
\hfill
\subfloat[The s-curve.]{\includegraphics[trim={3cm 2cm 2cm 4cm}, clip, scale=0.4]{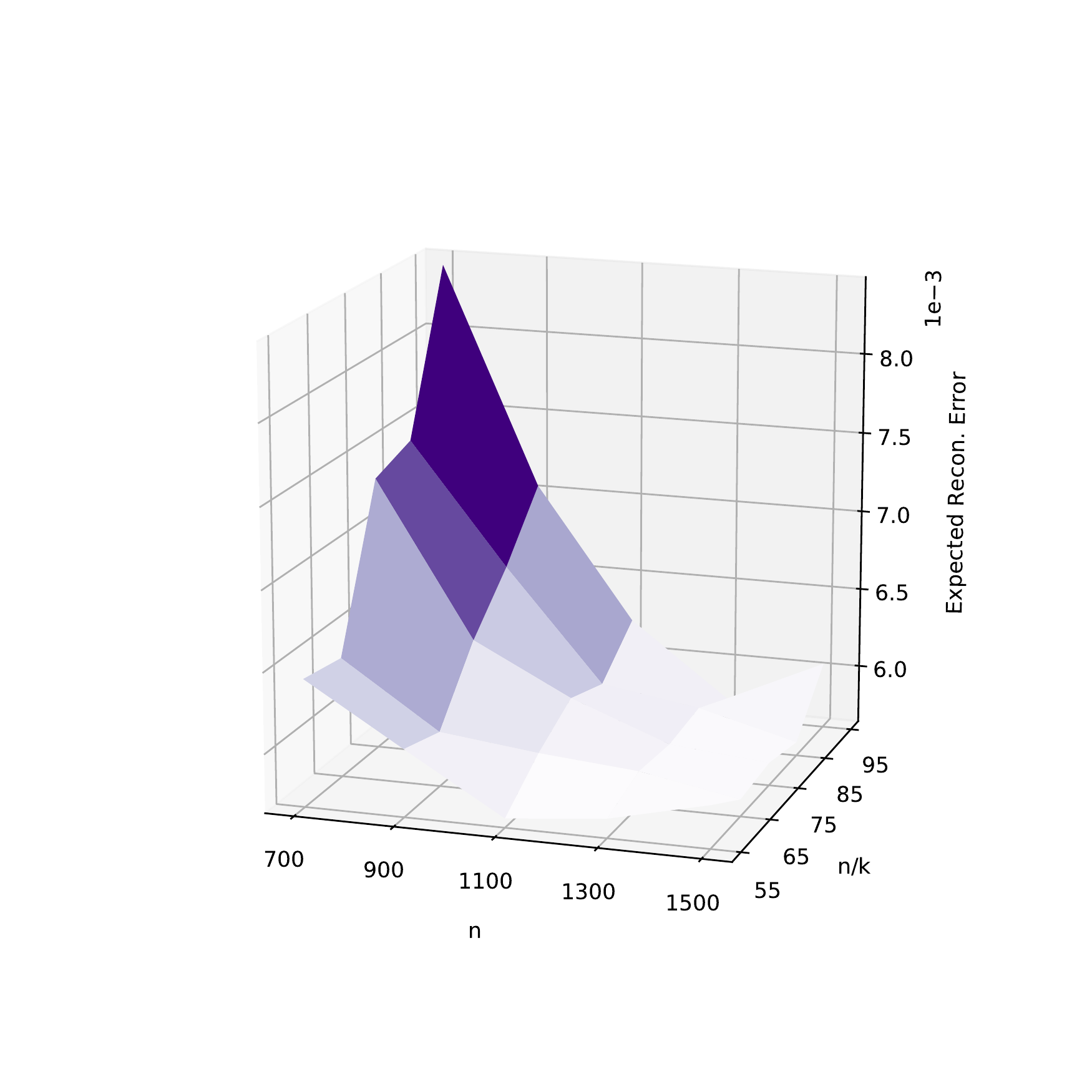}}
\hfill
\subfloat[The swiss roll.]{\includegraphics[trim={4cm 2cm 2cm 4cm}, clip, scale=0.4]{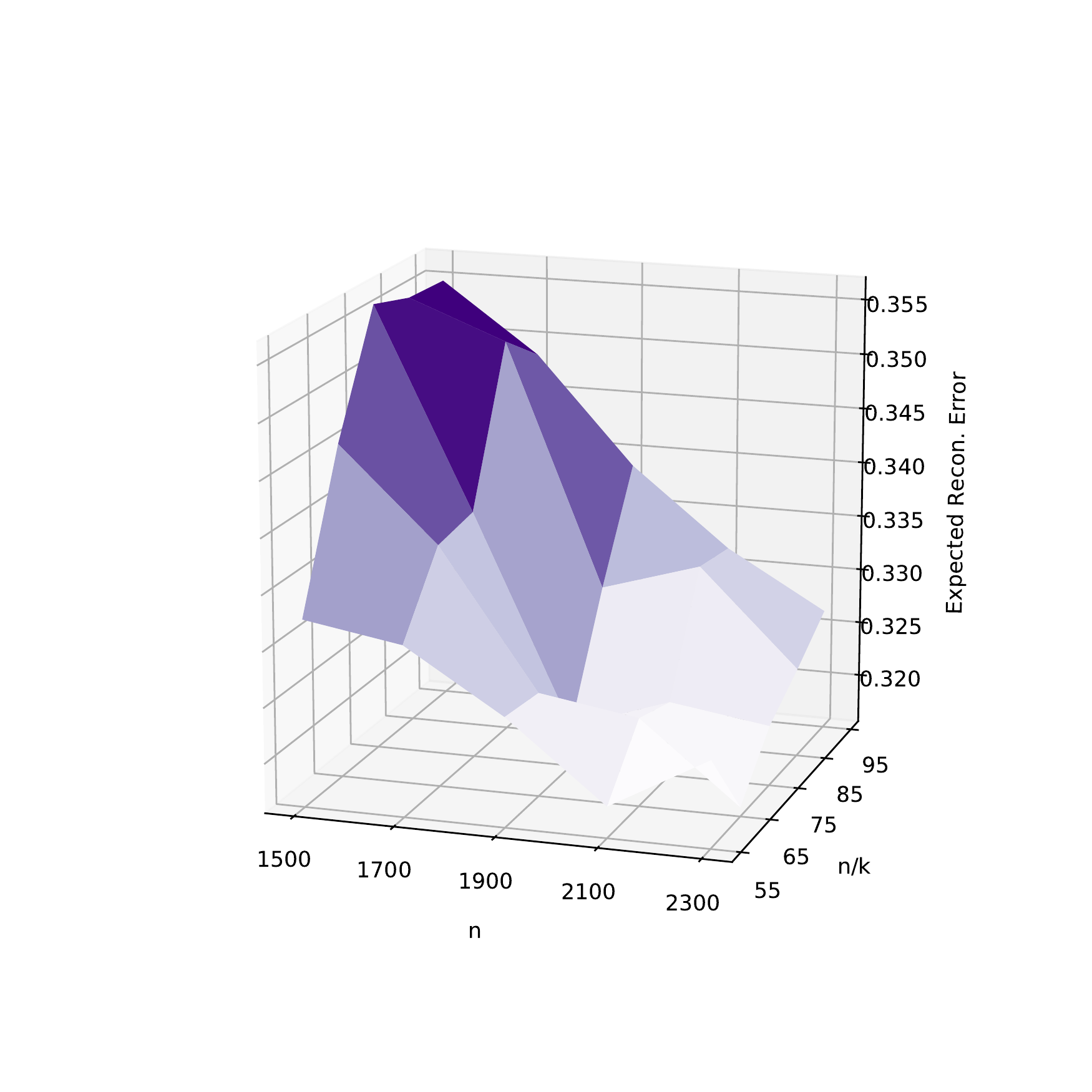}}
\caption{Expected reconstruction error landscape as a function of $n$ and $n/k$ for M2.}
\label{fig:M2_ere}
\vspace{-15pt}
\end{figure*}

Upon varying $n/k$ and $n$, we would expect a general trend of estimation quality to improve as $n/k$ decreases (the number of tangent spaces $k$ increases) and as the training set size $n$ increases. Figure \ref{fig:M2_ere} represents the landscape of expected reconstruction error as these quantities vary for the best estimate over 10 repetitions for M2 (the best performing method overall), and the general trend is as expected for all three (M1 and M3 not shown). Since expected reconstruction error is more robust to outliers, we omit the corresponding Hausdorff plots but the trend is the same, albeit noisier.

\subsection{MNIST Experiments}
The simulated datasets used above help shed light on the behavior of each of the methods and are often used in manifold research.  However, they are simulated and low dimensional and therefore not reflective of many real world datasets. Thus, we ran M2 on MNIST \citep{lecun-mnisthandwrittendigit-2010} with a latent dimensionality of 6 and randomly selected a neighborhood to explore.

\begin{figure}[ht]
\centering
\includegraphics[width=\linewidth]{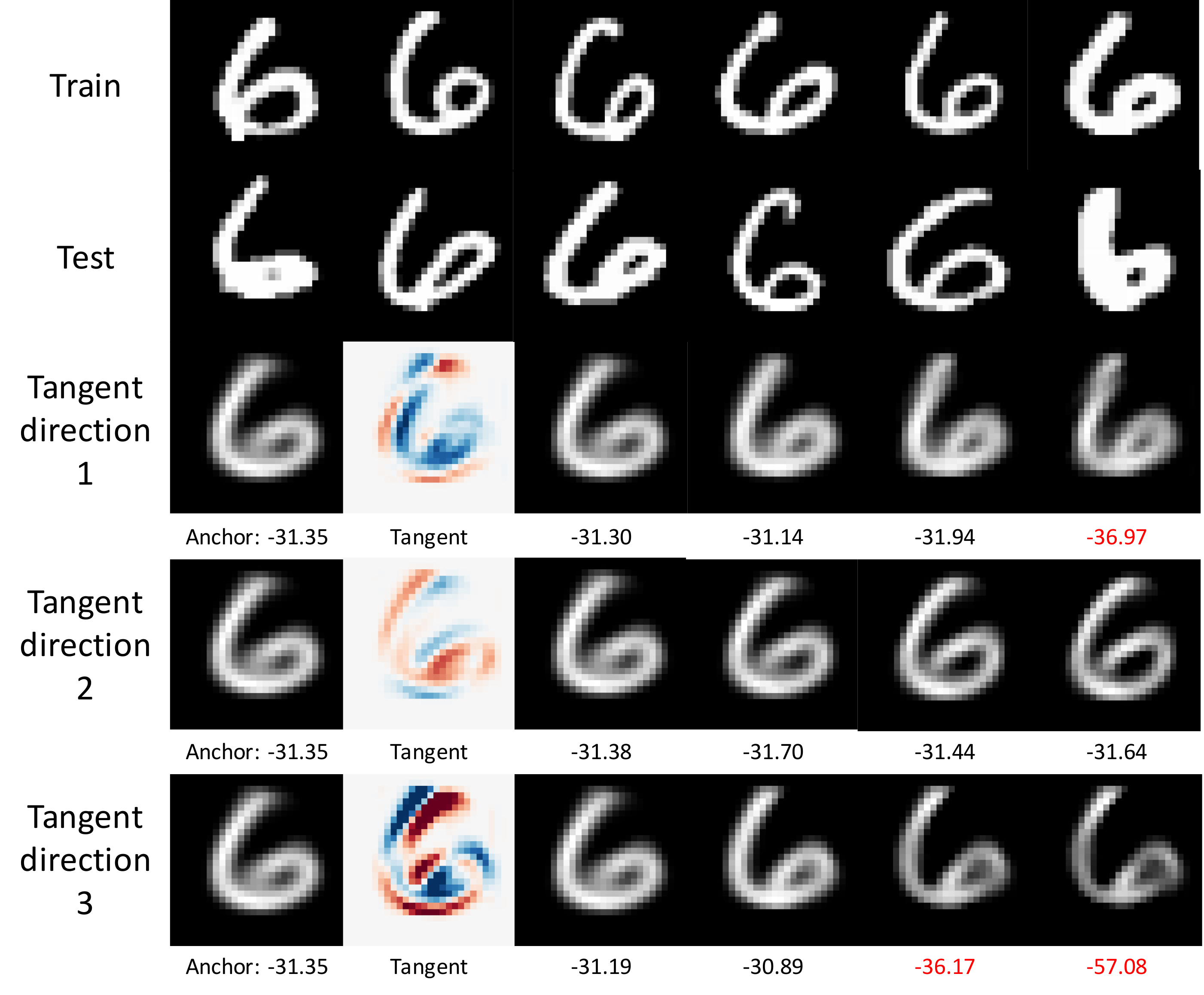}
\caption{Results from a random neighborhood. From top row down, random training images, some random test images, then for three tangent directions, anchor point, tangent (in RdBu colormap) and result of taking four steps in that direction from the tangent point along with the densities under the latent GMM. Density text in red denotes that the image is outside the faithful neighborhood density threshold. Gifs provided in supplementary material.}
\label{fig:mnist}
\end{figure}

In this neighborhood, we look at the anchor point and tangent directions.  From here, we take steps in each tangent direction from the anchor point and look at the density model to see when we leave the faithful neighborhood estimated.  Figure \ref{fig:mnist} presents the results. We believe that each directional derivative is altering the anchor in a meaningful way. For example, in row three, the directional derivative is making the step of the six more vertical and the loop tighter. This shows that there is diversity and interpretability for each of the directional derivatives. In addition, observe how the densities change as we move in each tangent direction.  For instance, in the last row the last two images which are splotchy fall outside the faithful neighborhood and the image found with density -30.89 is the highest scoring image under the density model and we argue, perceptually the most realistic to a handwritten digit.

\textbf{Supplementary material and reproducibility:} Supplementary plots and tables for Hausdorff distance and dataset configurations not shown here can be found at \url{https://www.dropbox.com/sh/mqsey3mbbgjccun/AABYHylHXMafrrXvV9zS2lY6a?dl=0}. In the spirit of reproducible research, the source code has been made available at \url{https://www.dropbox.com/sh/n5f91civznimgwr/AACdiFkz5d_BPV2jaeT5bJXFa?dl=0}.

\section{Conclusions}
\label{sect:concl}
In this paper, we formulated three methods of utilizing tangent bundle learners for the purpose of manifold estimation based upon the idea of faithful neighborhood estimation.  FNE assumes each tangent space to be a faithful linear approximator to the manifold in some neighborhood and then uses the points from $\mathcal{X}$ assigned to that tangent space to estimate the neighborhood over which it is a faithful approximator. The overall manifold estimate is then simply the union of these faithful neighborhoods. We studied the behavior of expected reconstruction error and Hausdorff distance between large uniform samples from the true and estimated manifolds as the number of training points $n$ and the ratio of the number of training points over the number of components to the tangent bundle learner $n/k$ varied.

This work is a step towards building a better understanding of how to make effective inferences in high dimensional spaces without running into the curse of dimensionality. It aims to highlight some aspects which might benefit inference tasks subsequent to manifold estimation/embedding. For example, classification algorithms based on manifold learning that use smallest distance to a sub-manifold from a test point (the asymmetric version of expected reconstruction error, $\mathcal{\varepsilon}_{\rho}$) could be implicitly assuming a poor manifold estimate:  one that is a superset of the true manifold, such as the ``shards of glass'' estimates shown in Figure \ref{fig:naive-lpca}. We contend that using FNE as a strategy to generalize better will improve performance of inference tasks, and will be the focus of our future work.

Finally, we should note that the idea of FNE is relevant for other approaches to manifold estimation that do not rely on a tangent bundle learner. For example, the method of \cite{silva2003geometric} uses thin-plate splines in each neighborhood centered at their anchor points as a local estimate of the manifold. While the approach allows for more flexibility in manifold estimation itself, the spline estimates still extend infinitely and faithful neighborhood estimation is relevant for all the reasons we have described in this work.

\bibliographystyle{plainnat}
\bibliography{Tiny,Ben,extras}

\end{document}